\pdfoutput=1
\documentclass[final,twocolumn]{svjour3}

\usepackage{xspace,amssymb,amsmath,graphicx,srcltx,dblfloatfix,url}
\usepackage{amsmath,amscd}
\usepackage{graphicx}
\graphicspath{{./}{./Fig/}{./Figs/}}

\usepackage{subfig}

\usepackage{natbib}
\bibpunct{(}{)}{;}{a}{}{;}

\usepackage{algorithm2e}
\let\savedalgorithm\algorithm
\let\savedendalgorithm\endalgorithm

\usepackage{algorithm}

\usepackage{color}

\usepackage[misc,geometry]{ifsym}

\usepackage{hyperref}  
\hypersetup{backref,colorlinks=true,linkcolor=blue,citecolor=blue,urlcolor=blue} 
\usepackage{etoolbox}
\makeatletter
\patchcmd{\NAT@citex}
  {\@citea\NAT@hyper@{\NAT@nmfmt{\NAT@nm}\NAT@date}}
  {\@citea\NAT@nmfmt{\NAT@nm}\NAT@hyper@{\NAT@date}}
  {}
  {}

\patchcmd{\NAT@citex}
  {\@citea\NAT@hyper@{%
     \NAT@nmfmt{\NAT@nm}%
     \hyper@natlinkbreak{\NAT@aysep\NAT@spacechar}{\@citeb\@extra@b@citeb}%
     \NAT@date}}
  {\@citea\NAT@nmfmt{\NAT@nm}%
   \NAT@aysep\NAT@spacechar%
   \NAT@hyper@{\NAT@date}}
  {}
  {}

\patchcmd{\NAT@citex}
  {\@citea\NAT@hyper@{%
     \NAT@nmfmt{\NAT@nm}%
     \hyper@natlinkbreak{\NAT@spacechar\NAT@@open\if*#1*\else#1\NAT@spacechar\fi}%
       {\@citeb\@extra@b@citeb}%
     \NAT@date}}
  {\@citea\NAT@nmfmt{\NAT@nm}%
   \NAT@spacechar\NAT@@open\if*#1*\else#1\NAT@spacechar\fi%
   \NAT@hyper@{\NAT@date}}
  {}
  {}
\makeatother

\renewcommand{\u}{\mathbf u}
\renewcommand{\v}{\mathbf v}

\newcommand{\x}{\mathbf x}
\newcommand{\w}{\mathbf w}

\newcommand{\z}{\mathbf z}

\newcommand{\e}{\boldsymbol e}

\newcommand{\A}{\mathbf A}
\newcommand{\C}{\mathbf C}
\newcommand{\X}{\mathbf X}
\renewcommand{\H}{\mathbf H}

\newcommand{\bmu}{{\boldsymbol \mu}}

\newcommand{\bSigma}{ {\boldsymbol \Sigma } }
\newcommand{\brho}{ {\boldsymbol \rho} }

\def\Phi{ { \phi } }

\newcommand{\Q}{\mathbf Q}
\newcommand{\q}{\mathbf q}

\newcommand{\eg}{{\em e.g.}\xspace}
\newcommand{\ie}{{\em i.e.}\xspace}

\newcommand{\st}{{\rm s.t.}\xspace}

\newcommand{\comment}[1]{}

\def\psd{\succcurlyeq}
\def\nsd{\preccurlyeq}

\newcommand{\cH}{\mathcal H}
\newcommand{\cX}{\mathcal X}

\def\T{{\!\top}}
\def\Real{\mathbb{R}}

\def\deltap{{\tilde{\delta}}}

\def\cite{\citep}

\newcommand{\norm}[2][2]{\ensuremath{ \left\| #2 \right\|_{ \mathrm{#1} } } }

\newcommand{\ADot}{ \ensuremath{  - \,} }

\newcommand{\modification}[1]{{\color{black}#1}}

\journalname{\it appearing in Int. J. Comput.
Vis.; 
content may change prior to final publication.}
\date{December  2012}

\begin{document}

\title{
      Training Effective Node Classifiers for Cascade Classification
}

\author{
        Chunhua Shen 
            \and 
        Peng~Wang 
            \and 
        Sakrapee~Paisitkriangkrai 
            \and 
        Anton~van~den~Hengel
        }
\institute{C. Shen (\Letter) 
                \and 
           P. Wang
                \and 
           S. Paisitkriangkrai
                \and
           A. van den Hengel
    \at
    Australian Centre for Visual Technologies, and
    School of Computer Science, The University of Adelaide, SA 5005, Australia
    \\
    \email{\url{chunhua.shen@adelaide.edu.au}}
%
%
    \\
    This work was in part supported by Australian Research Council 
    Future Fellowship FT120100969.
}

\maketitle

\begin{abstract}

         Cascade classifiers are widely used in real-time object detection.
         Different from conventional classifiers that are designed for a low overall classification
         error rate, a classifier in each node of the cascade is required to achieve an
         extremely high detection rate and moderate false positive rate.  Although there are a few
         reported methods addressing  this requirement in the context of object detection, there is
         no principled feature selection method that explicitly takes into account this asymmetric
         node learning objective.  We provide such an algorithm here.
         We show that a special case of the
         biased minimax probability machine has the same formulation as the linear asymmetric
         classifier (LAC) of  \citet{wu2005linear}. We then  design a new boosting algorithm that
         directly optimizes the cost function of LAC. The resulting totally-corrective boosting
         algorithm is implemented by the column generation technique in convex optimization.
         Experimental results on object detection verify the effectiveness of the proposed boosting
         algorithm as a node classifier in cascade object detection, and show performance better
         than that of the current state-of-the-art.

\keywords{
         AdaBoost
           \and
         Minimax Probability Machine
           \and
         Cascade Classifier
           \and
         Object Detection
           \and 
         Human Detection
        }

\end{abstract}

\section{Introduction}
\label{sec:intro}

        Real-time object detection inherently involves searching a large number of candidate image
        regions for a small number of objects.  Processing a single image, for example, can require
        the interrogation of well over a million scanned windows in order to uncover a single
        correct detection.  This imbalance in the data has an impact on the way that detectors are
        applied, but also on the training process.  This impact is reflected in the need to identify
        discriminative features from within a large over-complete feature set.
		
        Cascade classifiers have been proposed as a potential solution to the problem of imbalance
        in the data \cite{viola2004robust,bi2006Comp,dundar2007,brubaker2008,wu2008fast}, and have
        received significant attention due to their speed and accuracy. In this work, we propose a
        principled method by which to train a {\em boosting}-based cascade of classifiers.

        The boosting-based cascade approach to object detection was introduced by
		Viola and Jones \cite{viola2004robust,Viola2002Fast},
        and has received significant subsequent attention 
		\cite{li2004float,pham07,pham08multi,paul2008fast,shen2008face,paul2009cvpr}.
        It also underpins the current state-of-the-art \cite{wu2005linear,wu2008fast}.

\begin{figure*}[t]
             \begin{center}
                 \includegraphics[width=0.6\textwidth]{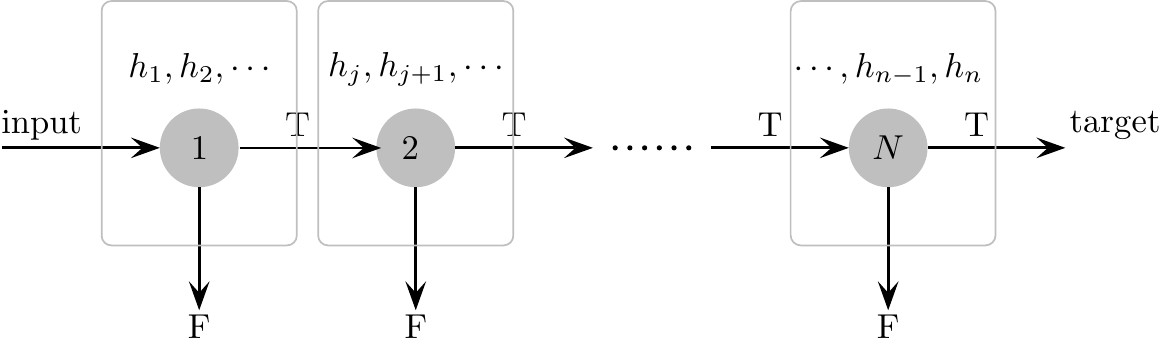}
                 \includegraphics[width=0.6\textwidth]{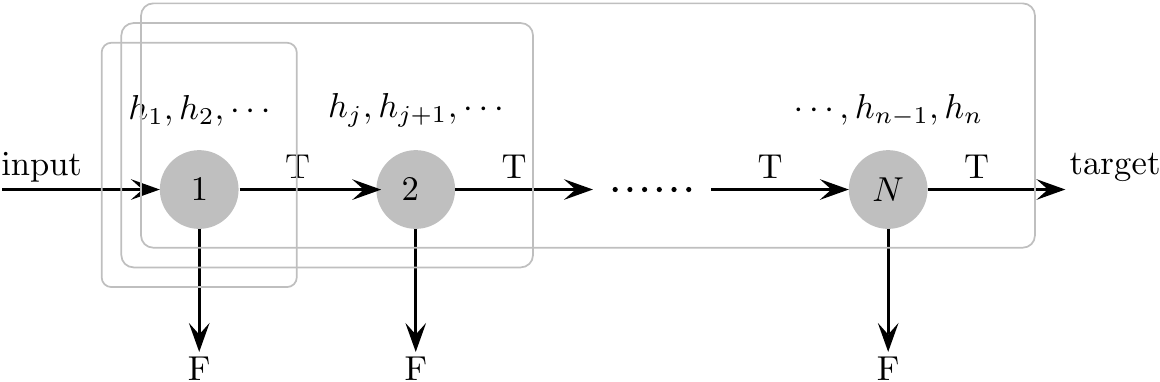}
             \end{center}
             \caption{Cascade classifiers.
             The first one is the standard cascade of \citet{viola2004robust}.
             The second one is the multi-exit cascade proposed in \citet{pham08multi}.
             Only those classified as true detection by all nodes will be true targets.}
             \label{fig:1A}
         \end{figure*}

        The Viola and Jones approach uses a cascade of increasingly complex classifiers, each of
        which aims to achieve the best possible classification accuracy while achieving an extremely
        low false negative rate.  These classifiers can be seen as forming the nodes of a degenerate
        binary tree (see Fig.~\ref{fig:1A}) whereby a negative result from any single such node
        classifier terminates the interrogation of the current patch.  Viola and Jones use AdaBoost
        to train each node classifier in order to achieve the best possible classification accuracy.
        A low false negative rate is achieved by subsequently adjusting the decision threshold until
        the desired false negative rate is achieved.  This process cannot be guaranteed to produce
        the best detection performance for a given false negative rate.

        Under the assumption that each node of the cascade classifier makes independent
        classification errors, the detection rate and false positive rate of the entire cascade are:
        $ F_{\rm dr} = \prod_{ t =1}^N d_t    $  and $ F_{\rm fp} = \prod_{ t =1}^N f_t    $,
        respectively, where $d_t$ represents the detection rate of classifier $t$, $f_t$ the
        corresponding false positive rate and $N$ the number of nodes.  As pointed out in 
		\cite{viola2004robust,wu2005linear}, these two equations suggest a {\em node learning
        objective}: Each node should have an extremely high detection rate $d_t $ ({\em e.g.},
        $99.7\%$) and a moderate false positive rate $ f_t $ ({\em e.g.}, $50\%$).  With the above
        values of $  d_t $ and $ f_t $, and a cascade of $ N = 20 $ nodes, then $ F_{\rm dr} \approx
        94\%$ and $ F_{\rm fp} \approx  10^ {-6} $, which is a typical design goal.

        One drawback of the standard AdaBoost approach to boosting
        is that it does not take advantage of the cascade classifier's special structure.
        AdaBoost only minimizes the overall classification error and
        does not  particularly  minimize the number of
        false negatives.  In this sense, the features selected by
        AdaBoost are not optimal for the purpose of
        rejecting as many negative examples as possible.
        Viola and Jones proposed a solution to this problem in AsymBoost \cite{Viola2002Fast}
        (and its variants \cite{pham07,pham08multi,wang2010asym,masnadi2007asym}) by
        modifying the  loss function so as to more greatly penalize false negatives.
        AsymBoost achieves better detection rates than AdaBoost, but still addresses the node
        learning goal \emph{indirectly}, and cannot be guaranteed to achieve the optimal solution.

        Wu et al.\ explicitly studied the node learning goal and proposed
        to use linear asymmetric classifier (LAC) and Fisher linear discriminant analysis (LDA)
        to adjust the weights on a set of features selected by AdaBoost or AsymBoost
        \cite{wu2005linear,wu2008fast}.
        Their experiments indicated that with this post-processing technique
        the node learning objective can be better met, which is translated into improved
        detection rates.
        In Viola and Jones' framework, boosting is used to select features and at the same time to
        train a strong classifier. Wu {et al.}'s work separates these two tasks:
        AdaBoost or AsymBoost is used to select features; and as a second step, LAC or LDA is used to
        construct a strong classifier by adjusting the weights of the selected features.
        The node learning objective is only considered at the
        second step. At the first step---feature selection---the node learning objective is
        not explicitly considered at all. We conjecture  that
        {\em further improvement may be gained
        if the node learning objective is explicitly
        taken into account at both steps}.
        We thus propose new  boosting algorithms to implement this idea and verify this conjecture.
        A preliminary version of this work was published in
        \citet{shen2010eccv}.

        Our major contributions are as follows.
        \begin{enumerate}
           \item
                Starting from the theory of minimax probability machines (MPMs),
                we derive a simplified
                version of the biased minimax probability machine, which has the same formulation as
                the linear asymmetric classifier of \citet{wu2005linear}.
                We thus show the underlying connection between MPM and LAC.
                Importantly, this new interpretation weakens some of the restrictions
                on the acceptable input data distribution imposed by LAC.
           \item
                We develop new boosting-like algorithms by directly
                minimizing the objective function of the
                linear asymmetric classifier, which results in an algorithm that we label LACBoost.
                We also propose FisherBoost on the basis of Fisher LDA rather than LAC.  Both
                methods may be used to identify the feature set that optimally achieves the node
                learning goal when training a cascade classifier.  To our knowledge, this is the
                first attempt to design such a feature selection method.
           \item
                LACBoost and FisherBoost share similarities with LPBoost
                \cite{Demiriz2002LPBoost} in the sense that both use
                column ge\-ne\-ra\-tion---a technique originally proposed
                for large-scale linear programming (LP).
                Typically, the Lagrange dual problem
                is solved at each iteration in column generation. We instead solve
                the primal qua\-dra\-tic programming (QP) problem, which has a special structure
                and entropic gradient (EG)
                can be used to solve the problem very efficiently.
                Compared with general interior-point based QP solvers, EG is much faster.
           \item
                We apply LACBoost and FisherBoost to object detection
                and better performance is observed over 
                o\-th\-e\-r methods \cite{wu2005linear,wu2008fast,maji2008}.
                In particular on pedestrian detection,
                FisherBoost achi\-eves the state-of-the-art, comparing with
                methods listed in \cite{Dollar2012Pedestrian} on three
                benchmark datasets.
                The results confirm our conjecture and show the
                effectiveness of LACBoost and FisherBoost.  These
                methods can be immediately applied to other asymmetric
                classification problems.
         \end{enumerate}

                Moreover, we analyze the condition that makes the validity of LAC,
                and show that the multi-exit cascade might be more suitable
                for applying LAC learning of \citet{wu2005linear}
                and \citet{wu2008fast} (and our LACBoost)
                rather than Viola-Jones' conventional cascade.

                As observed in \citet{wu2008fast}, in many cases, LDA
                even performs better than LAC. In our experiments, we
                have also observed similar phenomena.
                \citet{paul2009cvpr} empirically showed that LDA's
                criterion can be used to achieve better detection
                results.  An explanation of why LDA works so well for
                object detection is missing in the literature.  Here
                we  demonstrate that in the context of object
                detection, LDA can be seen as a regularized version of
                LAC in approximation.

                The proposed LACBoost/FisherBoost algorithm differs
                from traditional boosting algorithms in that it does
                not minimize a loss function.  This opens new
                possibilities for designing  boosting-like  algorithms
                for special purposes.  We have also extended column
                generation for optimizing nonlinear optimization
                problems.
                Next we review related work in the context of
                real-time object detection using cascade classifiers. 

        \subsection{Related Work}

        The field of object detection has made a significant progress over
        the last decade, especially after the seminal work of
        Viola and Jones.
        Three key components that contribute to their first robust {\em real-time}
        object detection framework are:
        \begin{enumerate}
        \item
              The cascade classifier, which efficiently filters out negative patches
              in early nodes while maintaining a very high detection rate;
        \item
            AdaBoost that selects informative features and at the same time trains
            a strong classifier;
        \item
            The use of integral images, which makes the computation of Haar features
            extremely fast.
        \end{enumerate}
        This approach has received significant subsequent attention.
        A number of alternative cascades have been developed including the
        soft cascade \cite{Bourdev05SoftCascade},
        WaldBoost \cite{Sochman2005WaldBoost},
        the dynamic cascade \cite{Rong2007},
        the AND-OR cascade \cite{dundar2007},
        the multi-exit cascade \cite{pham08multi},
        the joint cascade \cite{Lefakis2010Joint} and recently proposed,
        the rate constraint embedded cascade (RCECBoost) \cite{Saberian2012Learning}.
        In this work we have adopted the multi-exit cascade of Pham {et al.}
        due to its effectiveness and efficiency as demonstrated in \citet{pham08multi}.
        The multi-exit cascade improves classification
        performance by using the results of all of the weak classifiers applied to a patch so far in
        reaching a decision at each node of the tree (see Fig.~\ref{fig:1A}).
        Thus the  $ n $-th node classifier uses the results of the weak classifiers associated with
        node $n$, but also those associated with the previous $n-1$ node classifiers in the cascade.
        We show below that LAC post-processing can enhance the multi-exit cascade, and that
        the multi-exit cascade more accurately fulfills the LAC requirement that the margin
        be drawn from a Gaussian distribution.

		In addition to improving the cascade structure,
        a number of improvements have been made on
        the learning algorithm for building node classifiers in a cascade.
		Wu {et al.}, for example, use fast forward feature selection to
        accelerate the training procedure \cite{wu2003rare}.
        \citet{wu2005linear} also showed that LAC may be used to deliver
        better classification performance.
        Pham and Cham recently proposed online asymmetric boosting
        that considerably reduces the training time required \cite{pham07}.
        By exploiting the feature
        statistics, 
        \citet{Pham2007Fast} have also designed a fast method to train weak classifiers.
        \citet{li2004float} proposed FloatBoost, which  discards
        redundant weak classifiers during AdaBoost's
        greedy selection procedure.
        \citet{Masnadi2011Cost} proposed cost-sensitive
        boosting algorithms which can be applied to different
        cost-sensitive losses by means of gradient descent.
        \citet{Liu2003KL} also proposed KLBoost, aiming to select features that maximize the
        projected Kullback-Leibler divergence and select feature weights by minimizing the
        classification error.
        Promising results have also been reported by
        LogitBoost \cite{Tuzel2008PAMI} that employs the logistic regression
        loss, and GentleBoost \cite{Torralba2007} that uses adaptive
        Newton steps to fit the additive model.
        Multi-instance boosting has been introduced to object detection 
		\cite{viola2005mil,dollar08mcl,lin2009mil},
        which does not require precisely labeled locations of the targets in
        training data.

        New features have also been designed for improving the detection
        performance. Viola and Jones'
        Haar features are not sufficiently discriminative for detecting
        more complex objects like pedestrians, or multi-view faces.
        Covariance features \cite{Tuzel2008PAMI} and histogram of oriented gradients (HOG)
        \cite{Dalal2005HOG} have been proposed in this context,
        and efficient implementation approaches (along the lines of integral images) are
        developed for each. Shape context, which can also exploit integral images
        \cite{aldavert2010integral}, was applied to human detection in thermal images
        \cite{wang2010thermal}.
        The local binary pattern (LBP) descriptor and its variants have  been shown
        promising performance on human detection \cite{mu2008lbp,zheng2010cslbp}.
        Recently, effort has been spent on combining complementary features, including:
        simple concatenation of HOG and LBP  \cite{wang2009hoglbp}, combination of
        heterogeneous local features in a boosted cascade
        classifier \cite{wu2008tradeoff},
        and Bayesian integration of intensity, depth and motion features
        in a mixture-of-experts model \cite{enzweiler2010multi}.

        The rest of the paper is organized as follows. We briefly review the concept of
        minimax probability machine and derive the new simplified version of
        biased minimax probability machine in Section \ref{sec:mpm}.
        Linear asymmetric classification and its connection
        to the minimax probability machine is discussed in Section \ref{sec:LAC}.
        In Section \ref{sec:LACBoost},
        we show how to design new boosting algorithms (LACBoost and FisherBoost)
        by rewriting the optimization
        formulations of LAC and Fisher LDA.
        The new boosting algorithms are applied to object detection
        in Section \ref{sec:exp} and we
        conclude the paper in Section \ref{sec:con}.

        \subsection{Notation}

        The following notation is used.
        A matrix is denoted by a bold upper-case
        letter ($\X$); a column vector is denoted by a bold lower-case
        letter ($ \x $).
        The $ i$th row of $\X $ is denoted by $ \X_{i:} $
        and the $ i $th column $ \X_{:i}$.
        The identity matrix is $ \bf I $ and its size should be clear
        from the context. $ \bf 1 $ and
        $ \bf 0 $ are column vectors of $ 1$'s and $ 0$'s,
        respectively.
        We use $ \psd, \nsd $ to denote component-wise inequalities.

        Let $ {\cal T} = \{ (\x_i, y_i  ) \}_{i = 1, \cdots, m}$ be the set of
        training data, where $ \x_i \in \cX$ and $ y_i \in \{-1,+1\}
        $, $ \forall i$.
        The training set consists of $ m_1 $ positive training points
        and $ m_2 $ negative ones; $ m_1 + m_2 = m $.
        Let $ h ( \cdot  ) \in \cH $ be a weak
        classifier that projects an input vector $ \x $ into
        $\{-1, +1 \}$.
        Note that here we consider only classifiers with discrete outputs
        although the developed methods can use real-valued
        weak classifiers too.
        We assume that $ \cH $,
        the set from which $ h ( \cdot  ) $ is selected, is finite and has
        $n$ elements.

        Define the matrix $ \H^{\cal Z} \in \Real^{ m \times n }$ such that the  $ (i,j)$ entry $
        \H^{\cal Z}_{ij} =  h_j ( \x_i ) $ is the label predicted by weak classifier $ h_j(\cdot) $ for
        the datum $ \x_i $, where $\x_i$ the $i$th element of the set ${\cal Z}$.
        In order to simplify
        the notation we eliminate the superscript when ${\cal Z}$ is the training set,
        so  $\H^{\cal Z} = \H$.
        Therefore, each
        column $ \H_{ :j }  $ of the matrix $ \H $ consists of the
        output of weak classifier $ h_j(\cdot) $ on all the training
        data; while each row $ \H_{ i: } $ contains  the outputs of all
        weak classifiers on the training datum $ \x_i $.
        Define similarly the matrix $ \A \in  \Real^{ m \times n }$ such that
        $ \A_{ij} = y_i h_j (  \x_i ) $.
        Note that boosting algorithms entirely depends on the matrix $  \A  $ and
        do not directly interact with the training examples.
        Our following discussion will thus largely focus on the matrix $\A$.
        We write the vector obtained by multiplying a matrix $  \A  $
        with a vector
        $ \w $ as $ \A \w $ and its $i$th entry as $ (\A \w)_i$.
        If we let $\w$ represent the coefficients of a selected weak
        classifier then the margin of the training datum $ \x_i $ is
        $ \rho_i = \A_{ i :} \w = (\A \w)_i $
        and the vector of such margins for all of the training data is $\brho = \A \w$.

\section{Minimax Probability Machines}
\label{sec:mpm}

    Before we introduce our boosting algorithm, let us briefly 
    review the concept of minimax probability machines (MPM) 
    \cite{lanckriet2002mpm} first.

\subsection{Minimax Probability Classifiers}

    Let $ \x_1  \in \Real^n $ and  $ \x_2 \in \Real^n  $
    denote two random vectors
    drawn from two distributions with means and covariances
    $ ( \bmu_1, \bSigma_1 ) $
    and $ (  \bmu_2, \bSigma_2  ) $, respectively. 
    Here $ \bmu_1, \bmu_2 \in \Real^n  $ and 
    $ \bSigma_1, \bSigma_2 \in \Real^{ n \times n } $. 
    We define the class labels of $ \x_1 $ and
    $ \x_2 $  as $ +1 $ and $ -1 $, w.l.o.g.  
    The minimax probability machine (MPM)
    seeks a robust separation hyperplane that can separate the two classes of data with
    the maximal probability. The hyperplane can be expressed as $ \w ^\T \x = b  $ with 
    $ \w \in \Real^n \backslash \{\bf 0 \} $ and $ b \in \Real $. 
    The problem of identifying the optimal hyperplane may then
    be formulated as 
    \begin{align}
        \label{eq:1}
        \max_{\w,b,\gamma} \,\, \gamma \,\,\,\, \st \,   
        & \left[  \inf_{ \x_1 \sim ( \bmu_1, \bSigma_1 )  } \Pr \{ \w^\T \x_1 \geq b \}  \right]
                  \geq \gamma,
        \\
        & \left[  \inf_{ \x_2 \sim ( \bmu_2, \bSigma_2 )  } \Pr \{ \w^\T \x_2 \leq b \}  \right]
                  \geq \gamma.    \notag       
    \end{align}
    Here $ \gamma $ is the lower bound of the classification accuracy
    (or the worst-case accuracy) on test data.
    This problem can be transformed into a convex problem,
    more specifically a second-order cone
    program (SOCP) \cite{boyd2004convex} and thus can be solved efficiently 
	\cite{lanckriet2002mpm}.       
    
 \subsection{Biased Minimax Probability Machines}

    The formulation \eqref{eq:1} assumes that the classification problem is balanced.
    It attempts to achieve a high recognition accuracy, which assumes that
    the losses associated with all mis-classifications are identical. 
    However, in many applications this is not the case.
    
    \citet{huang2004mpm} proposed a biased version of MPM
    through a slight modification of \eqref{eq:1}, which
    may be formulated as 
    \begin{align}
        \label{eq:2}
        \max_{\w,b,\gamma} \,\, \gamma \,\,\,\, \st \,   
        & \left[  \inf_{ \x_1 \sim ( \bmu_1, \Sigma_1 )  } \Pr \{ \w^\T \x_1 \geq b \}  \right]
                  \geq \gamma,
        \\
         & \left[  \inf_{ \x_2 \sim ( \bmu_2, \Sigma_2 )  } \Pr \{ \w^\T \x_2 \leq b \}  \right]
                  \geq \gamma_\circ.    \notag       
    \end{align}
    Here $ \gamma_\circ \in (0,1) $ is a prescribed constant,
    which is the acceptable classification accuracy for the less
    important class.
    The resulting decision hyperplane 
    prioritizes the classification of
    the important class $ \x_1 $ 
    over that of the less important class $ \x_2 $.
    Biased MPM is thus expected to perform better in 
    biased classification applications.

    Huang et al. showed that \eqref{eq:2} can be iteratively
    solved via solving a sequence of SOCPs using 
    the fractional programming (FP) technique.
    Clearly it is
    significantly more computationally demanding
    to solve \eqref{eq:2} than \eqref{eq:1}.

    Next we show how to re-formulate \eqref{eq:2} into a simpler quadratic program (QP) 
    based on the recent theoretical results in \cite{yu2009general}. 

\subsection{Simplified Biased Minimax Probability Machines}
    
\modification{
   In this section, we are interested in simplifying the problem of \eqref{eq:2} 
   for a special case of $ \gamma_\circ = 0.5 $, due to its important
   application in object detection \cite{viola2004robust,wu2005linear}.
   In the following discussion, for simplicity,
   we only consider $ \gamma_\circ = 0.5 $ although some algorithms
   developed may also apply to $  \gamma_\circ < 0.5 $.      

   Theoretical results in \cite{yu2009general} show that,
   the worst-case constraint in \eqref{eq:2} can be written in different forms 
   when $\x$ follows arbitrary, symmetric, symmetric unimodal or Gaussian distributions 
   (see Appendix \ref{App:MPMa}).    
   Both the MPM \cite{lanckriet2002mpm} and the biased MPM \cite{huang2004mpm} are based the most
   general form of the four cases shown in Appendix \ref{App:MPMa}, 
   \ie, Equation \eqref{eq:5A}  for arbitrary distributions, 
   as they do not impose constraints upon the distributions of $\x_1$ and $\x_2$.
  
   However, one may take advantage of structural information whenever
   available. 
   For example, it is shown in \cite{wu2005linear} that, for the face detection problem, 
   weak classifier outputs can be well approximated by the Gaussian distribution. 
   In other words, the constraint for arbitrary distributions does not utilize 
   any  type of {\it a priori} information, 
   and hence, for many problems, considering arbitrary distributions
   for simplifying \eqref{eq:1} and \eqref{eq:2} is too {\em conservative}. 
   Since both the MPM \cite{lanckriet2002mpm} and the biased MPM \cite{huang2004mpm}
   do not assume any constraints on the distribution family,
   they fail to exploit this structural information.
  
   Let us consider the special case of $ \gamma_\circ = 0.5 $. 
   It is easy to see that the worst-case constraint in \eqref{eq:2}
   becomes a simple linear constraint for symmetric, symmetric unimodal,
   as well as Gaussian distributions (see Appendix \ref{App:MPMa}).  
   As pointed out in \cite{yu2009general}, such a result is the immediate 
   consequence of symmetry because the worst-case distributions are forced to
   put probability mass arbitrarily far away on both sides of the mean. 
   In such case, any information about the covariance is neglected.   

   We now apply this result to the biased MPM as represented by \eqref{eq:2}.
   Our main result is the following theorem. 
    \begin{theorem}
        With $ \gamma_\circ = 0.5 $, the biased minimax problem \eqref{eq:2} 
        can be formulated as an unconstrained problem:
        \begin{equation}
            \max_{\w  } \,\,
            \frac{ \w^\T ( \bmu_1 - \bmu_2 ) } 
                { \sqrt{ \w^\T \bSigma_1 \w } }, \label{eq:11d} 
        \end{equation}
        under the assumption
        that $ \x_2 $ follows a symmetric distribution.
        The optimal $b$ can be obtained through: 
        \begin{equation}
            b = \w^\T \bmu_2. \label{EQ:11c} 
        \end{equation}
        The worst-case classification accuracy for the first class, 
        $ \gamma^\star$, is obtained by solving
        \begin{equation}
            \label{eq:opt}
            \varphi( \gamma^\star )
            = \frac{ -b^\star + {a^\star } ^ \T \bmu_1 }
                   { 
                     \sqrt{ {\w^\star} ^\T \Sigma_1 \w^\star }
                   },
        \end{equation}
        where         
        \begin{equation}
            \label{eq:6}
            \varphi( \gamma ) = 
            \begin{cases}
                \sqrt{ \frac{ \gamma }{ 1 - \gamma } }           & \text{ ~ if~} 
                            \x_1 \sim ( \bmu_1, \Sigma_1 ),
                \\
                \sqrt{ \frac{1}{ 2 (1 - \gamma) }  }               & \text{ ~ if~} 
                            \x_1 \sim ( \bmu_1, \Sigma_1 )_{ \rm S},
                \\
                \frac{2}{3}\sqrt{ \frac{1}{ 2 (1 - \gamma) }  }    & \text{ ~ if~} 
                            \x_1 \sim ( \bmu_1, \Sigma_1 )_{ \rm SU},
                \\
                \Phi^{-1} ( \gamma )        & \text{ ~ if~} 
                                            \x_1 \sim {\cal G}( \bmu_1, \Sigma_1 ).
            \end{cases}
        \end{equation}
        and $\{ \w^\star , b^\star \}$ is the optimal solution of
        \eqref{eq:11d} and \eqref{EQ:11c}.
    \label{thm:2}
    \end{theorem}
    Please refer to Appendix \ref{App:MPMa} for the proof of Theorem \ref{thm:2}.

    We have derived the biased MPM algorithm from a different perspective. 
    We reveal that only the assumption of symmetric distributions is needed
    to arrive at a simple unconstrained formulation.
    Compared with the approach in \cite{huang2004mpm},
    we have used more information to simply the optimization problem. 
    More importantly, as will be shown in the next section, this unconstrained
    formulation enables us to design a new boosting algorithm. 

        There is a close connection between our algorithm 
        and the linear asymmetric classifier (LAC) in \cite{wu2005linear}.
        The resulting problem \eqref{eq:11d} is exactly the same as 
        LAC in \cite{wu2005linear}. 
        Removing the inequality in this constraint 
        leads to a problem solvable by eigen-decomposition.
        We have thus shown that the results of Wu et al. may be generalized from the Gaussian
        distributions assumed in \cite{wu2005linear} to symmetric distributions.
          
}

\section{Linear Asymmetric Classification}
\label{sec:LAC}

        We have shown that starting from the biased minimax probability
        machine, we are able to obtain the same optimization formulation
        as shown in \citet{wu2005linear}, while much weakening the
        underlying assumption (symmetric distributions versus Gaussian
        distributions).
        Before we propose our LACBoost and FisherBoost, however,
        we provide a brief overview of LAC.

        \citet{wu2008fast}
        proposed linear asymmetric classification
        (LAC) as a post-processing step
        for training nodes in the cascade framework.
        In \cite{wu2008fast}, it is stated that
        LAC is guaranteed to reach an optimal solution
        under the assumption of Gaussian data distributions.
        We now know that this Gaussianality
        condition may be relaxed.

        Suppose that we have a linear classifier
        \[
            f(\x) = {\bf sign}(\w^\T \x - b).
        \]
        We seek a  $\{ \w , b \}$ pair with a very
        high accuracy on the positive data $\x_1$ and a moderate accuracy on
        the negative $\x_2$.
        This can be expressed as the following problem:
    \begin{align}
        \max_{\w \neq {\bf 0}, b}
        &
        \,  \Pr_{\x_1 \sim ( \bmu_1, \bSigma_1) }
        \{ \w ^\T \x_1 \geq b \}, \,\,
        \notag
        \\
        {\rm s.t.}  & \, \Pr_{\x_2 \sim (\bmu_2,\bSigma_2) }
        \{ \w^\T \x_2
            \leq b \} = \lambda. 
    \label{eq:LAC}
    \end{align}
    In \cite{wu2005linear},
    $\lambda$ is set to $0.5$ and it is assumed that for any $\w$,
    $\w^\T \x_1$
    is Gaussian and $\w^\T \x_2$ is symmetric,
    \eqref{eq:LAC} can be approximated by
    \eqref{eq:11d}.
    Again, these assumptions may be relaxed as we have shown in the last section.
    Problem \eqref{eq:11d} is similar to LDA's optimization problem
    \begin{equation}
        \label{EQ:LDA1}
        \max_{\w \neq \bf 0} \;\;
        \frac{  \w^\T ( \bmu_1 -  \bmu_2  ) }
        {  \sqrt{  \w^\T ( \bSigma_1 + \bSigma_2 )  \w  }  }.
    \end{equation}
    Problem \eqref{eq:11d} can be solved by eigen-decomposition and a closed-form
    solution can be derived:
\begin{equation}
            \w^\star = \bSigma_1^{-1} ( \bmu_1 - \bmu_2 ),
    \quad
    b^{\star} = { \w^{\star} } ^{\T} \bmu_2.
\label{eq:11d_SOL}
\end{equation}
On the other hand, each node in cascaded boosting classifiers has the following form:
\begin{equation}
    \label{EQ:nodeclassifier}
    f(\x) = {\bf sign}(\w^\T \H (\x) - b).
\end{equation}
 We override the symbol $ \H (\x)$ here,
 which denotes the output vector of all weak classifiers over the datum $ \x $.
    We can cast each node as a linear classifier over the feature space
constructed by the binary outputs of all weak classifiers.
For each node in a cascade classifier, we wish to maximize the detection
rate while maintaining the false positive rate at a
moderate level  (for example, around $50.0\%$).
    That is to say,  the problem
    \eqref{eq:11d} represents the node learning goal.
    Boosting algorithms such as AdaBoost can be used as feature
    selection methods, and LAC is used to learn a linear classifier over
    those binary features chosen by boosting as in
    \citet{wu2005linear}.
    The advantage of this approach is that LAC considers the asymmetric
    node learning explicitly.

However, there is a precondition on the  validity of LAC that
for any $\w$, $\w^\T \x_1$ is a Gaussian and $\w^\T \x_2$
is symmetric.
In the case of boosting classifiers, $\w^\T \x_1$ and $\w^\T \x_2$ can be
expressed as the margin of positive data and negative data, respectively.
Empirically  
\citet{wu2008fast}
verified that $\w^\T \x$ is  approximately Gaussian for a cascade face detector.
We discuss this issue in more detail in Section~\ref{sec:exp}.
\citet{shen2010dual}
theoretically proved that
under the assumption that weak classifiers are independent,
the margin of AdaBoost follows the Gaussian distribution,
as long as the number of weak classifiers is {\em sufficiently large}.
In Section~\ref{sec:exp} we verify this theoretical result by performing the
normality test on nodes with different number of weak classifiers.

\section{Constructing Boosting Algorithms from LDA and LAC}
\label{sec:LACBoost}

    In kernel methods, the original data are non\-linear\-ly
    mapped to a feature space by a mapping
    function $ \Psi ( \cdot ) $. The function need not be known,
    however, as rather than being applied to the data directly,
    it acts instead through the inner product
    $ \Psi ( \x_i ) ^\T \Psi ( \x_j )  $.
    In boosting \cite{Ratsch2002BoostSVM}, however,
    the mapping function can be seen as being explicitly known,
    as
    $
        \Psi ( \x ) : \x \mapsto [ h_1(\x),\dots,h_n(\x) ].
    $
    Let us consider the Fisher LDA case first because the solution to LDA
    will generalize to LAC straightforwardly, by looking at
    the similarity between \eqref{eq:11d} and \eqref{EQ:LDA1}.

    Fisher LDA
    maximizes the between-class variance and minimizes the within-class
    variance. In the binary-class case, the more general formulation in
     \eqref{EQ:LDA1} can be expressed as
    \begin{equation}
        \label{EQ:100}
        \max_\w \;\;  \frac{ ( \bmu_1 - \bmu_2 ) ^ 2 }
                         { \sigma_1 + \sigma_2 }
            =
                    \frac{   \w ^\T   \C_b \w }
                         {   \w ^\T  \C_w \w },
    \end{equation}
    where $ \C_b $ and $ \C_w $ are the between-class and within-class
    scatter matrices; $ \bmu_1 $ and $ \bmu_2 $ are
    the projected centers of the two classes.
    The above problem can be equivalently reformulated as
    \begin{equation}
        \label{EQ:101}
        \min_\w \;\;  \w ^\T \C_w \w -  \theta ( \bmu_1 - \bmu_2  ),
    \end{equation}
    for some certain constant $ \theta $ and under the assumption that
    $ \bmu_1 - \bmu_2 \geq 0 $.\footnote{In our object detection experiment,
    we found that this assumption can always be satisfied.}
    Now in the feature space, our data are
    $ \Psi( \x_i ) $, $ i=1\dots m$.
    	Define the vectors $ \e, \e_1, \e_2 \in \Real^{m}$ such that $ \e = \e_1 + \e_2 $,
		the $ i $-th entry of $ \e_1 $ is $1/m_1$ if $ y_i = +1 $ and $0$ otherwise, and
		the $ i $-th entry of $ \e_2 $ is $1/m_2$ if $ y_i = -1 $ and $0$ otherwise.
    We then see that
    \begin{align}
        \bmu_1
        &  = \frac{ 1 } { m_1 } \w^\T \sum_{y_i = 1}  \Psi(\x_i)
           = \frac{ 1 } { m_1 } \sum_{y_i = 1} \A_{ i: } \w
           \notag
         \\
         &
        = \frac{ 1 } { m_1 } \sum_{y_i = 1} (\A \w)_i
           = \e_1 ^\T \A \w,
    \end{align}
    and
    \begin{align}
     \bmu_2
     & =
     \frac{ 1 } { m_2 }   \w^\T \sum_{y_i = -1}  \Psi(\x_i)
     = \frac{ 1 } { m_2 } \sum_{y_i = -1} \H_{ i: } \w
     = - \e_2 ^\T  \A \w,
    \end{align}
    For ease of exposition we order the training data according to their
        labels so
        the vector $ \e \in \Real^{m}$:
        \begin{equation}
        \e = [ 1/m_1,\cdots, 1/m_2,\cdots  ]^\T,
            \label{EQ:e}
        \end{equation}
        and the first $ m_1$ components of $ \brho $ correspond to the
        positive training data and the remaining ones
        correspond to the $ m_2$
        negative data.
        We now see that $ \bmu_1 - \bmu_2 = \e^\T \brho  $,
        $ \C_w =  {m_1 }/{ m } \cdot \bSigma_1 + {m_2 }/{ m } \cdot  \bSigma_2 $
        with
        $ \bSigma_{1,2} $ the covariance matrices.
        Noting that
        \[
        \w^\T \bSigma_{1,2} \w = \frac{1}{m_{1,2} ( m_{1,2} - 1 ) }
        \sum_{i>k, y_i=y_k = \pm 1}
        (\rho_i - \rho_k )^2,
        \]
        we can easily rewrite the original problem \eqref{EQ:100} (and \eqref{EQ:101})
        into:
        \begin{align}
            \min_{\w,\brho}
             ~&
            \tfrac{1}{2} \brho ^\T \Q \brho - \theta \e^\T
            \brho,
          \notag\\
            \quad {\rm s.t.} ~&\w \psd {\bf 0},
             {\bf 1}^\T \w = 1,
      \notag \\
      &
     {\rho}_i = ( \A \w )_i,
        i = 1,\cdots, m.
            \label{EQ:QP1}
        \end{align}
        Here
        $ \Q = \begin{bmatrix} \Q_1 & {\bf 0} \\ {\bf 0} & \Q_2  \end{bmatrix} $
        is a block matrix with
        \[
        \Q_1 =
        \begin{bmatrix}
                \tfrac{1}{m} & -\tfrac{1}{ m (m_1-1)} & \ldots & -\tfrac{1}{m(m_1-1)} \\
                -\tfrac{1}{m(m_1-1)} & \tfrac{1}{ m } & \ldots & -\tfrac{1}{m(m_1-1)} \\
                \vdots & \vdots & \ddots & \vdots \\
                -\tfrac{1}{m(m_1-1)} & -\tfrac{1}{m (m_1-1)} & \ldots &\tfrac{1}{m }
        \end{bmatrix},
        \]
        and $ \Q_2 $ is similarly defined by replacing $ m_1$ with $ m_2 $ in $ \Q_1$:
        \[
        \Q_2 =
            \begin{bmatrix}
                \tfrac{1}{m}         & -\tfrac{1}{m(m_2-1)} & \ldots &
                -\tfrac{1}{m(m_2-1)}                                \\
                -\tfrac{1}{m(m_2-1)} & \tfrac{1}{m}         & \ldots &
                -\tfrac{1}{m(m_2-1)}                                \\
                \vdots               & \vdots & \ddots & \vdots     \\
                -\tfrac{1}{m(m_2-1)} & -\tfrac{1}{m(m_2-1)} & \ldots
                &\tfrac{1}{m}
            \end{bmatrix}.
        \]
  Also note that we have introduced a constant $ \frac{1}{2} $ before the quadratic term
  for convenience. The normalization
                  constraint $ { \bf 1 } ^\T \w = 1$
                  removes the scale ambiguity of $ \w $. Without it the problem is
                  ill-posed.

  We see from the form of \eqref{eq:11d} that the covariance of
  the negative data is not involved in
  LAC and thus
  that if we set
   $ \Q = \begin{bmatrix} \Q_1 & {\bf 0} \\ {\bf 0} & \bf 0  \end{bmatrix} $ then
  \eqref{EQ:QP1} becomes the optimization problem of LAC.

  At this stage, it remains unclear about how to solve the problem \eqref{EQ:QP1}
  because we do not know all the weak classifiers.
        There may be extremely (or even infinitely) many weak
         classifiers in $ \cH $, the set from which
        $ h ( \cdot  ) $ is selected, meaning that the dimension of the optimization
        variable $ \w $ may also be extremely large.
    So \eqref{EQ:QP1} is a semi-infinite quadratic program (SIQP).
    We show how column generation can be used to solve this problem.
    To make column generation applicable, we need to derive a
    specific Lagrange dual of the primal
    problem.

\subsection{The Lagrange Dual Problem}

\label{sec:dual_prob}

    We now derive the Lagrange dual of the quadratic problem \eqref{EQ:QP1}.
    Although we are only interested in the variable $ \w $, we need to
    keep the auxiliary variable $ \boldsymbol  \rho $ in order to obtain
    a meaningful dual problem. The Lagrangian of \eqref{EQ:QP1}
    is
    \begin{align}
        L (
           \underbrace{ \w, \brho}_{\rm primal}, \underbrace{ \u, r }_{\rm dual}
        )
    & = \tfrac{1}{2} \brho ^\T \Q \brho   -  \theta \e^\T \brho
    + \u ^\T ( \brho - \A \w  ) - \q ^\T \w
    \notag
    \\
    & + r ( {\bf 1} ^\T \w - 1 ),
    \label{EQ:Lag1}
    \end{align}
    with
     $
     \q \psd \bf 0
     $.
    $ \sup_{\u, r} \inf_{ \w, \brho } L ( \w, {\brho}, \u, r  ) $
    gives the following  Lagrange dual:
           \begin{align}
               \max_{\u, r} ~& -r - \overbrace{
                       \tfrac{1}{2}
                      (\u - \theta \e)^\T \Q^{-1} (\u - \theta \e)
                      }^{\rm regularization},
                      \notag\\
            {\rm \;\; s.t.}
                     ~
                      &
                      \sum_{i=1}^m u_i \A_{i:} \nsd r {\bf 1 } ^\T.
            \label{EQ:dual}
        \end{align}
        In our case, $ \Q $ is rank-deficient and its inverse does not exist
        (for both LDA and LAC).
        Actually for both $ \Q_1 $ and $ \Q_2 $, they have a zero
        eigenvalue with the corresponding eigenvector being all ones.
        This is easy to see because for $ \Q_1 $ and $ \Q_2 $,
        the sum of each row (or each column) is zero. 
        We can simply regularize $ \Q $ with $ \Q + \deltap {\bf I} $ with
        $ \deltap $ a  small positive constant.
        Actually, $ \Q $ is a diagonally dominant matrix but not strict diagonal dominance.
        So $ \Q + \deltap {\bf I} $ with any $ \deltap > 0 $ is strict
        diagonal dominance and by the Gershgorin circle theorem,
        a strictly diagonally   dominant matrix must be invertible.

        One of the  KKT optimality conditions between the dual and
        primal is
        \begin{equation}
         \brho^\star = - \Q^{-1} ( \u ^ \star - \theta \e ),
        \end{equation}
        which can be used to establish the connection between the dual optimum and
        the primal optimum.
        This is obtained by the fact that
        the gradient of $ L $ w.r.t. $ \brho $ must vanish at
        the optimum, $ { \partial L } / { \partial \rho_i } = 0  $,
        $ \forall i = 1\cdots n $.

        Problem \eqref{EQ:dual} can be viewed as a regularized LPBoost problem.
        Compared with the hard-margin LPBoost \cite{Demiriz2002LPBoost},
        the only difference is the regularization term in the cost function.
        The duality gap between the primal \eqref{EQ:QP1} and the
        dual \eqref{EQ:dual} is zero. In other words, the solutions of
        \eqref{EQ:QP1} and \eqref{EQ:dual} coincide.
        Instead of solving \eqref{EQ:QP1} directly, one calculates the
        most violated constraint in \eqref{EQ:dual} iteratively for
        the current solution and adds this constraint to the
        optimization problem.  In theory, any column that violates
        dual feasibility can be added.  To speed up the convergence,
        we add the most violated constraint by solving the following
        problem:
      \begin{equation}
          h' ( \cdot ) =  {\rm argmax}_{h( \cdot ) } ~
            \sum_{i=1}^m u_i y_i h ( \x_i).
         \label{EQ:pickweak}
      \end{equation}
        This  is exactly the same as the one that standard AdaBoost
        and LPBoost use for producing the best weak classifier at each iteration. 
		That is to say, to find the weak classifier that has the minimum weighted
        training error.  We summarize the LACBoost/FisherBoost
        algorithm in
        Algorithm~\ref{alg:QPCG}.
        By simply changing  $ \Q_2 $, Algorithm~\ref{alg:QPCG} can be used to
        train either LACBoost or FisherBoost.
        Note that to obtain an actual strong classifier,
        one may need to include an offset $ b $, {\em i.e.} the final classifier
        is $ \sum_{j=1}^n h_j (\x) - b $ because from the cost function
        of our algorithm \eqref{EQ:101}, we can see that the cost function itself
        does not minimize any classification error. It only finds a projection
        direction in which the data can be maximally separated. A simple line
        search can find an optimal $ b $.
        Moreover, when training a cascade, we need to tune this offset anyway
        as shown in \eqref{EQ:nodeclassifier}.

        The convergence of Algorithm~\ref{alg:QPCG} is guaranteed by
        general column generation or cutting-plane algorithms, which
        is easy
        to establish:
        \begin{theorem}
            \label{thm:CG} 
             The column generation procedure decreases the
             objective value of problem \eqref{EQ:QP1} 
             at each iteration and hence in the limit
             it solves the problem \eqref{EQ:QP1}  globally to a desired
            accuracy. 
        \end{theorem}
        The proof is deferred to Appendix \ref{App:CG}.
        In short, when a new $ h'(\cdot) $ that violates dual
        feasibility is added, the new optimal value of the dual
        problem (maximization) would decrease.  Accordingly, the
        optimal value of its primal problem decreases too because they
        have the same optimal value due to zero duality gap. Moreover
        the primal cost function is convex, therefore in the end it
        converges to the global minimum.

   \linesnumbered\SetVline
   \begin{algorithm}[t]
   \caption{Column generation for SIQP.}
   \centering
   {\small
   \begin{minipage}[]{.94\linewidth}
   \KwIn{Labeled training data $(\x_i, y_i), i = 1\cdots m$;
         termination threshold $ \varepsilon > 0$;
         regularization
         parameter $ \theta $; maximum number of iterations
         $ n_{\rm max}$.
    }
       { {\bf Initialization}:
            $ m = 0 $;
            $ \w = {\bf 0} $;
            and $ u_i = \frac{1}{ m }$, $ i = 1$$\cdots$$m$.
   }

   \For{ $ \mathrm{iteration} = 1 : n_\mathrm{max}$}
   {
     \ADot
         Check for the optimality: \\
         {\bf if}{ $ \mathrm{iteration}  > 1 $ \text{ and } $
                    \sum_{ i=1 }^m  u_i y_i h' ( \x_i )
                           < r + \varepsilon $},
                  \\
                  { \bf then}
                  \\
                  $~ ~ ~$ break;  and the problem is solved;

     \ADot
         Add $  h'(\cdot) $ to the restricted master problem, which
         corresponds to a new constraint in the dual;

      \ADot
         Solve the dual problem \eqref{EQ:dual}
         (or the primal problem \eqref{EQ:QP1})
         and update $ r $ and
         $ u_i$ ($ i = 1\cdots m$).

      \ADot
         Increment the number of weak classifiers
             $n = n + 1$.
   }
   \KwOut{
         The selected features are $ h_1, h_2, \dots, h_n $.
         The final strong classifier is:
         $ F ( \x ) = \textstyle \sum_{j=1}^{ n } w_j h_j( \x ) - b $.
         Here the offset $ b $ can be learned by a simple line search.
   }
   \end{minipage}
   } 
   \label{alg:QPCG}
   \end{algorithm}

    At each iteration of column generation,
    in theory, we can  solve either the dual \eqref{EQ:dual}
    or the primal problem \eqref{EQ:QP1}.
    Here we choose to solve an equivalent variant of the primal problem \eqref{EQ:QP1}:
     \begin{align}
            \min_{\w} ~& \tfrac{1}{2} \w ^\T ( \A^\T \Q \A) \w
            - ( \theta \e ^\T
            \A ) \w,
            \;\;
            {\rm s.t.}
            \;
            \w \in \Delta_n,
            \label{EQ:QP2}
     \end{align}
	where $ \Delta_n$ is the unit simplex, 
	which is defined as $ \{ \w \in \Real^n  :  {\bf 1 } ^ \T \w = 1, \w \psd {\bf 0 } \} $. 

    In practice, it could be much faster to solve \eqref{EQ:QP2} since
    \begin{enumerate}
    \item
        Generally,
        the primal problem has a smaller size, hence faster to solve.
        The number of variables of \eqref{EQ:dual} is $ m $ at each iteration,
        while the number of variables is the number of iterations
        for the primal problem.
        For example, in Viola-Jones' face detection framework,
        the number of training data $ m =
        10,000 $ and  $ n_{\rm max} = 200 $. In other words, the
        primal problem has at most $ 200 $ variables in this case;

    \item
        The dual problem \eqref{EQ:dual} is a standard QP problem. It
        has no special structure to exploit. As we will show, the
        primal problem \eqref{EQ:QP2} belongs to
        a special class of problems and can be efficiently
        solved using entropic/\-ex\-p\-o\-ne\-n\-t\-i\-a\-t\-e\-d
        gradient descent (EG) \cite{beck03mirror,globerson07exp}.
        See Appendix \ref{App:EG} for details of the EG algorithm.

        A fast QP solver is extremely important for training our
        object detector since we need to  solve a few thousand
        QP problems.
        Compared with standard QP solvers like Mosek \cite{Mosek},
        EG is much faster. EG makes it possible
        to train a detector using almost the same amount of time
        as using standard AdaBoost because the majority of time is
        spent on weak classifier training and bootstrapping.
     \end{enumerate}

    We can recover both of the dual variables
    $ \u^\star, r^\star $ easily from
    the primal variable $ \w^\star, \brho^\star $:
    \begin{align}
        \u^\star &=  - \Q\brho^\star  + \theta \e; \label{EQ:KA}\\
         r^\star  &=   \max_{ j = 1 \dots n }
         \bigl\{ \textstyle \sum_{i=1}^m u_i^\star \A_{ij} \bigr\}.
         \label{EQ:KAb}
    \end{align}
    The second equation is obtained by the fact that
     in the dual problem's  constraints, at optimum,
    there must exist at least one  $ u_i^\star$
    such that the equality holds. That is to say,
    $ r^\star $ is the largest {\em edge}
    over all weak classifiers.

        In summary, when using EG to solve the primal problem,
        Line $ 5 $ of Algorithm~\ref{alg:QPCG} is:

        \ADot
        {\em Solve the primal problem \eqref{EQ:QP2} using EG, and update
        the dual variables $ \u $ with \eqref{EQ:KA}, and $ r $ with \eqref{EQ:KAb}.
        }

\section{Experiments}
\label{sec:exp}

    In this section, we
    perform our experiments on both synthetic and challenging
    real-world data sets, \eg, face and pedestrian detection.

\subsection{Synthetic Testing}

\begin{figure}[t!]
    \centering
        \includegraphics[width=.45\textwidth]{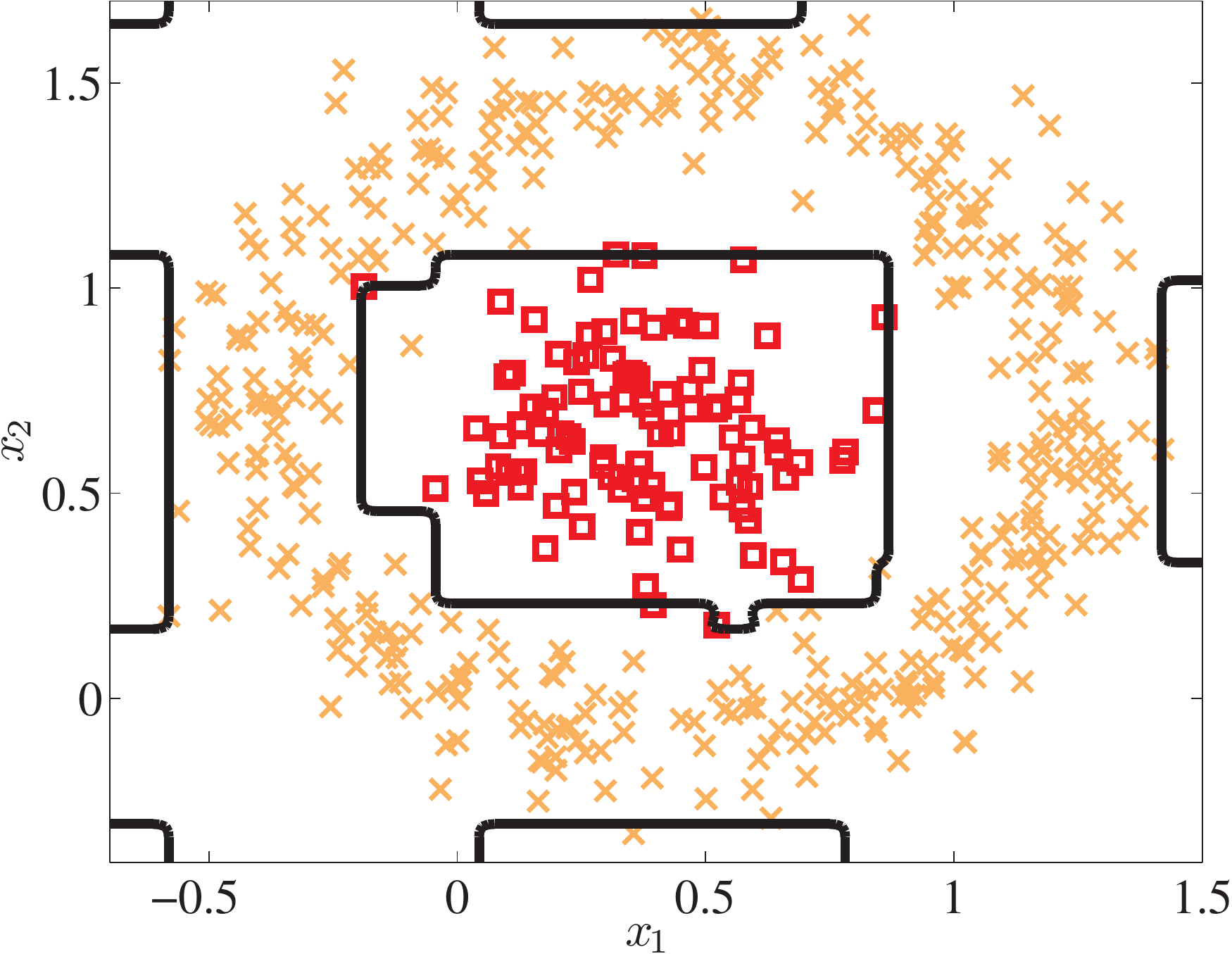}
        \includegraphics[width=.45\textwidth]{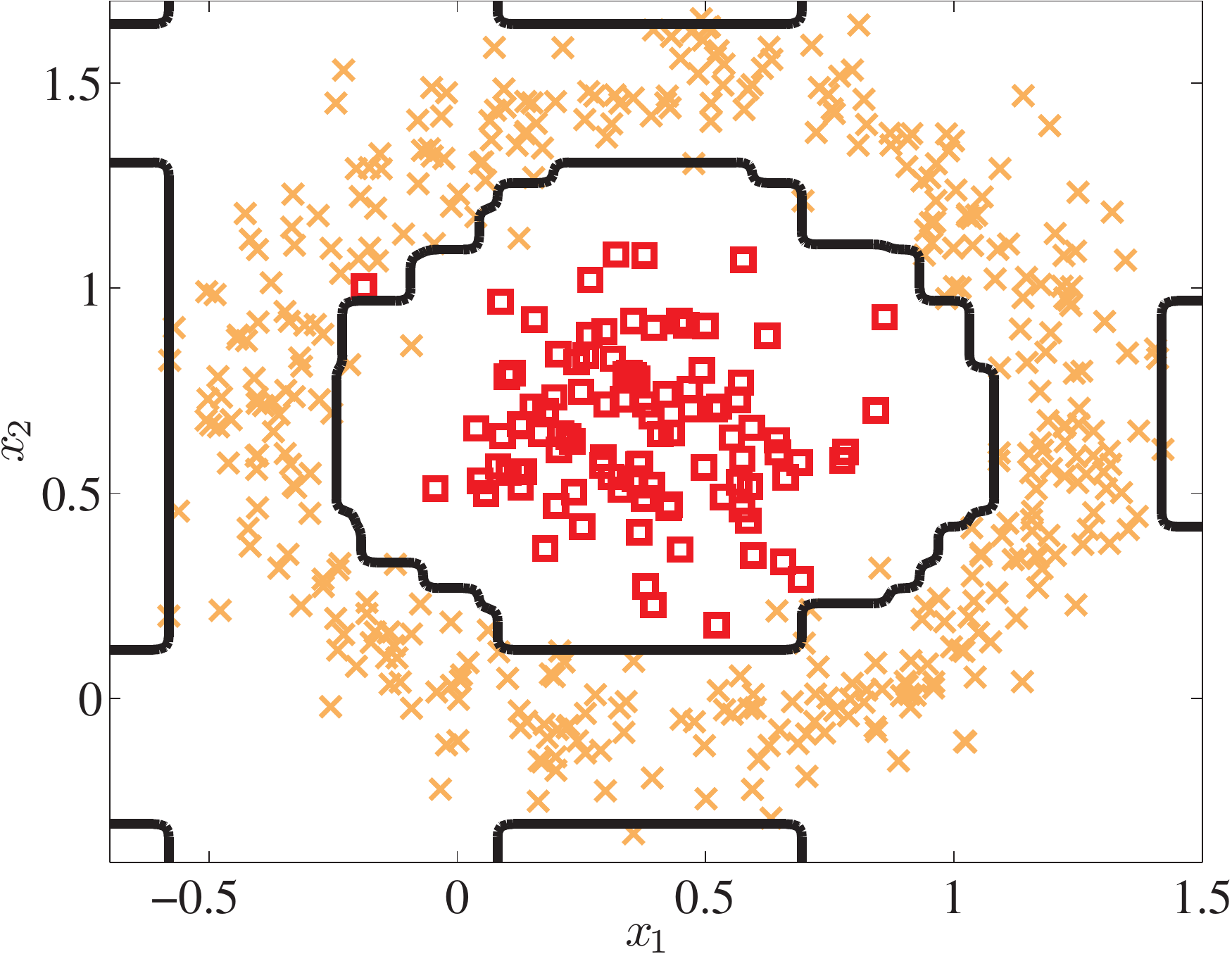}
    \caption{Decision boundaries of
    AdaBoost (top) and FisherBoost (bottom) on $2$D artificial data
    generated from the Gaussian distribution 
    (positive data represented by $ \square $'s and negative data
    by $\times$'s). 
    Weak classifiers are vertical and horizontal decision stumps.
    FisherBoost emphasizes more on positive samples than negative samples.
    As a result, the decision boundary of FisherBoost is more similar to the
    Gaussian distribution than the decision boundary of AdaBoost.
    }
    \label{fig:toy}
\end{figure}

    We first illustrate the performance of FisherBoost on an asymmetrical
    synthetic data set where there are a large number of negative samples
    compared to the positive ones.
    Fig. \ref{fig:toy} demonstrates the subtle difference
    in classification boundaries between AdaBoost and FisherBoost.
    It can be observed that
    FisherBoost places more emphasis on positive samples than negative samples
    to ensure these positive samples would be classified correctly.
    AdaBoost, on the other hand, treat both positive and negative samples equally.
    This might be due to the fact that AdaBoost only optimizes the overall
    classification accuracy.
    This finding is consistent with our results reported earlier in \cite{paul2009cvpr, GSLDA2010Shen}.


\begin{table*}[t!]
  \begin{center}
  \scalebox{0.8}
  {
  \begin{tabular}{r|ccccccccc}
  \hline
          & AdaBoost &  LAC & FLDA & AsymBoost & CS-ADA & RCBoost$^{1}$  & RCBoost$^{2}$  & LACBoost & FisherBoost \\
  \hline
  \hline
Digits      & $99.30$ ($0.10$)  & $99.30$ ($0.21$) & $99.37$ ($0.08$) & $\mathbf{99.40}$ ($\mathbf{0.11}$)  & $99.37$ ($0.09$) & $99.36$ ($0.17$) & $99.27$ ($0.15$) & $99.12$ ($0.07$) & $\mathbf{99.40}$ ($\mathbf{0.13}$) \\
Faces       & $98.70$ ($0.14$)  & $98.78$ ($0.42$) & $98.86$ ($0.22$) & $98.73$ ($0.14$) & $98.71$ ($0.20$) & $98.75$ ($0.18$) & $98.66$ ($0.23$) & $98.63$ ($0.29$) & $\mathbf{98.89}$ ($\mathbf{0.15}$) \\
Cars        & $97.02$ ($1.55$)  & $97.07$ ($1.34$) & $97.02$ ($1.50$) & $97.11$ ($1.36$) & $97.47$ ($1.31$) & $96.84$ ($0.87$) & $96.62$ ($1.08$) & $96.80$ ($1.47$) & $\mathbf{97.78}$ ($\mathbf{1.27}$) \\
Pedestrians & $98.54$ ($0.34$)  & $98.59$ ($0.71$) & $98.69$ ($0.28$) & $98.55$ ($0.45$) & $98.51$ ($0.36$) & $98.67$ ($0.29$) & $98.65$ ($0.39$) & $\mathbf{99.12}$ ($\mathbf{0.35}$)  & $98.73$ ($0.33$) \\
Scenes      & $99.59$ ($0.10$)  & $99.54$ ($0.21$) & $99.57$ ($0.12$) & $99.66$ ($0.12$) & $99.68$ ($0.10$) & $99.61$ ($0.19$) & $99.62$ ($0.16$) & $97.50$ ($1.07$) & $\mathbf{99.66}$ ($\mathbf{0.10}$) \\
\hline
Average     & $98.63$  & $98.66$ & $98.70$ & $98.69$ & $98.75$ & $98.64$ & $98.56$ & $98.23$ & $\mathbf{98.89}$ \\
  \hline
  \end{tabular}
  }
  \end{center}
  \caption{
  \modification{
  Test errors (\%) on five real-world data sets.
  All experiments are run $5$ times with $100$ boosting iterations.
  The average detection rate and standard deviation (in percentage) at $50\%$ false positives are reported.
  Best average detection rate is shown in boldface.
  }}
  \label{tab:responsesA1}
\end{table*}

{

\subsection{Comparison With Other Asymmetric Boosting}

In this experiment, FisherBoost and LACBoost are compared against
several asymmetric boosting algorithms, namely,
AdaBoost with LAC or Fisher LDA post\--pro\-ce\-ssing \cite{wu2008fast}, AsymBoost \cite{Viola2002Fast}, cost-sensitive AdaBoost (CS-ADA) \cite{Masnadi2011Cost}
and rate constrained boo\-st\-ing (RCBoost)
\cite{Saberian2012Learning}. The results of AdaBoost are also
presented as the baseline.
For each algorithm, we train a strong classifier consisting of $100$ weak classifiers along with their coefficients.
The threshold was determined such that the false positive rate of test set is $50\%$. For every method, the experiment is repeated $5$ times and the average detection rate on positive class is reported.
For FisherBoost and LACBoost, the parameter $\theta$ is chosen from $\{1/10,$ $1/12,$ $1/15,$ $1/20\}$ by cross\--va\-li\-da\-tion.
For AsymBoost, we choose $k$ (asymmetric factor) from $\{ 2^{0.1},$ $2^{0.2},$ $\cdots,$ $2^{0.5} \}$ by cross\--va\-li\-da\-tion.
For CS-ADA, we set the cost for misclassifying positive and negative data as follows.
We assign the asymmetric factor $k = C_1/C_2$ and restrict $0.5(C_1 + C_2) = 1$.
We choose $k$ from $\{1.2,$ $1.65,$ $2.1,$ $2.55,$ $3\}$ by cross-\-va\-li\-da\-tion.
For RCBoost, we conduct two experiments.
In the first experiment, we use the same training set to enforce the target detection rate, while in the second experiment; we use $75\%$ of the training data to train the model and the other $25\%$ to enforce the target detection rate.
We set the target detection rate, $D_T$, to $99.5\%$, the barrier coefficient, $\gamma$, to $2$ and the number of iterations before halving $\gamma$, $N_d$, to $10$.

We tested the performance of all algorithms on five real-world data sets,
including both machine learning (USPS) and vision data sets (cars, faces, pedestrians, scenes).
We categorized USPS data sets into two classes: even digits and odd
digits.
For faces, we use face data sets from \cite{viola2004robust} and randomly extract $5000$ negative patches from background images.
We apply principle component analysis (PCA) to preserve $95\%$ total variation.
The new data set has a dimension of $93$.
For UIUC car \cite{Agarwal2004Learning}, we downsize the original image from $40 \times 100$ pixels to $20 \times 50$ pixels and apply PCA.
The projected data capture $95\%$ total variation and has a final dimension of $228$.
For Daimler-Chrysler pedestrian data sets \cite{Munder2006Experimental}, we apply PCA to the original $18 \times 36$ pixels.
The projected data capture $95\%$ variation and has a final dimension of $139$.
For indoor/outdoor scene, we divide the $15$-scene data set used in \cite{Lazebnik2006Beyond} into $2$ groups: indoor and outdoor scenes.
We use CENTRIST as our feature descriptors and build $50$ visual code words using the histogram intersection kernel \cite{Wu2011CENTRIST}.
Each image is represented in a spatial hierarchy manner.
Each image consists of $31$ sub-windows.
In total, there are $1550$ feature dimensions per image.
All $5$ classifiers are trained to remove $50\%$ of the negative data, while retaining almost all positive data.
We compare their detection rates in Table~\ref{tab:responsesA1}.
From our experiments, FisherBoost demonstrates the best performance on most data sets.
However, LACBoost does not perform as well as expected.
We suspect that the poor performance might partially due to 
numerical issues, which can cause overfitting. 
We will discuss this in more detail in Section \ref{sec:why}. 

}

\subsection{Face Detection Using a Cascade Classifier}

\begin{figure*}[tbp!]
    \centering
        \includegraphics[width=.32\textwidth]{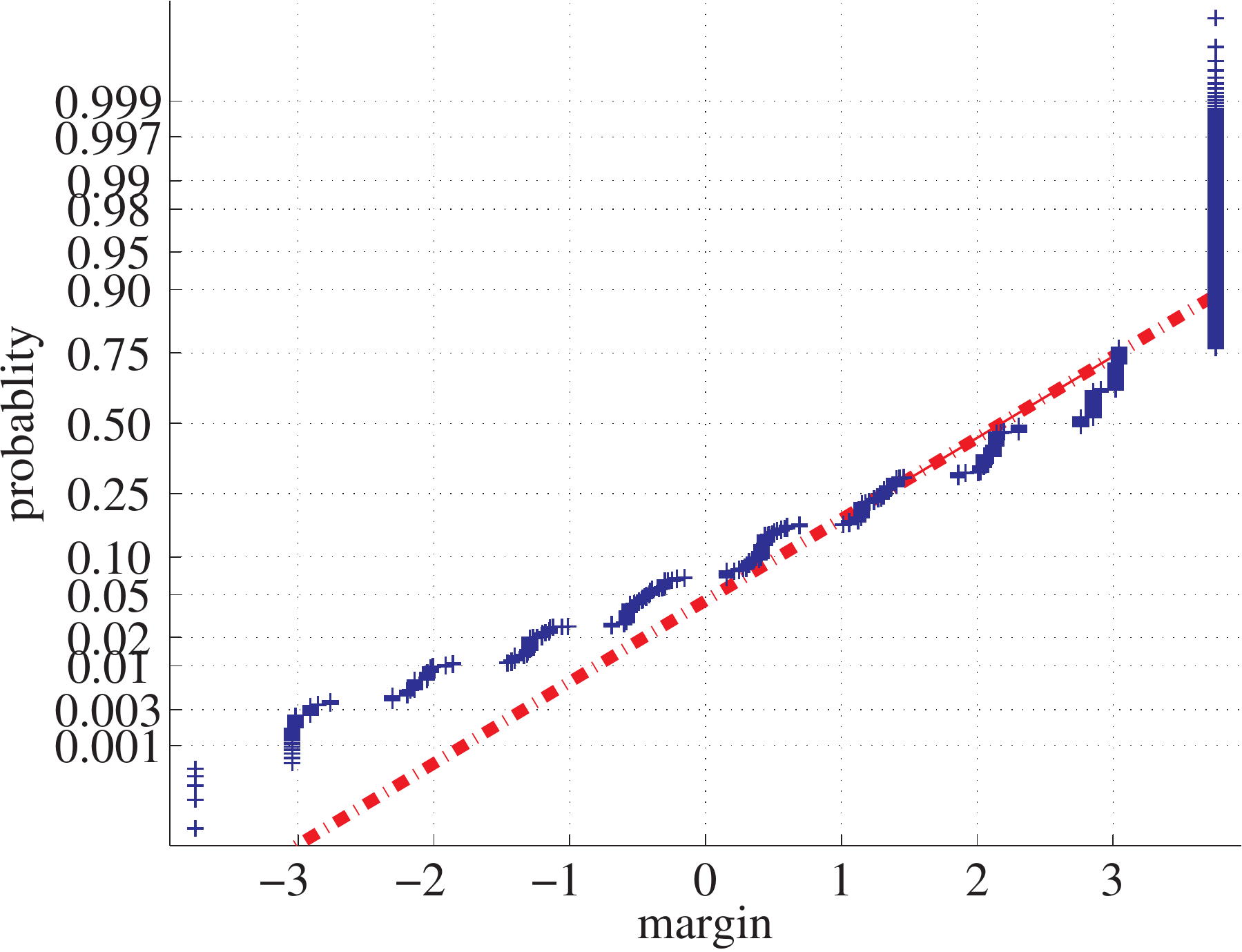}
        \includegraphics[width=.32\textwidth]{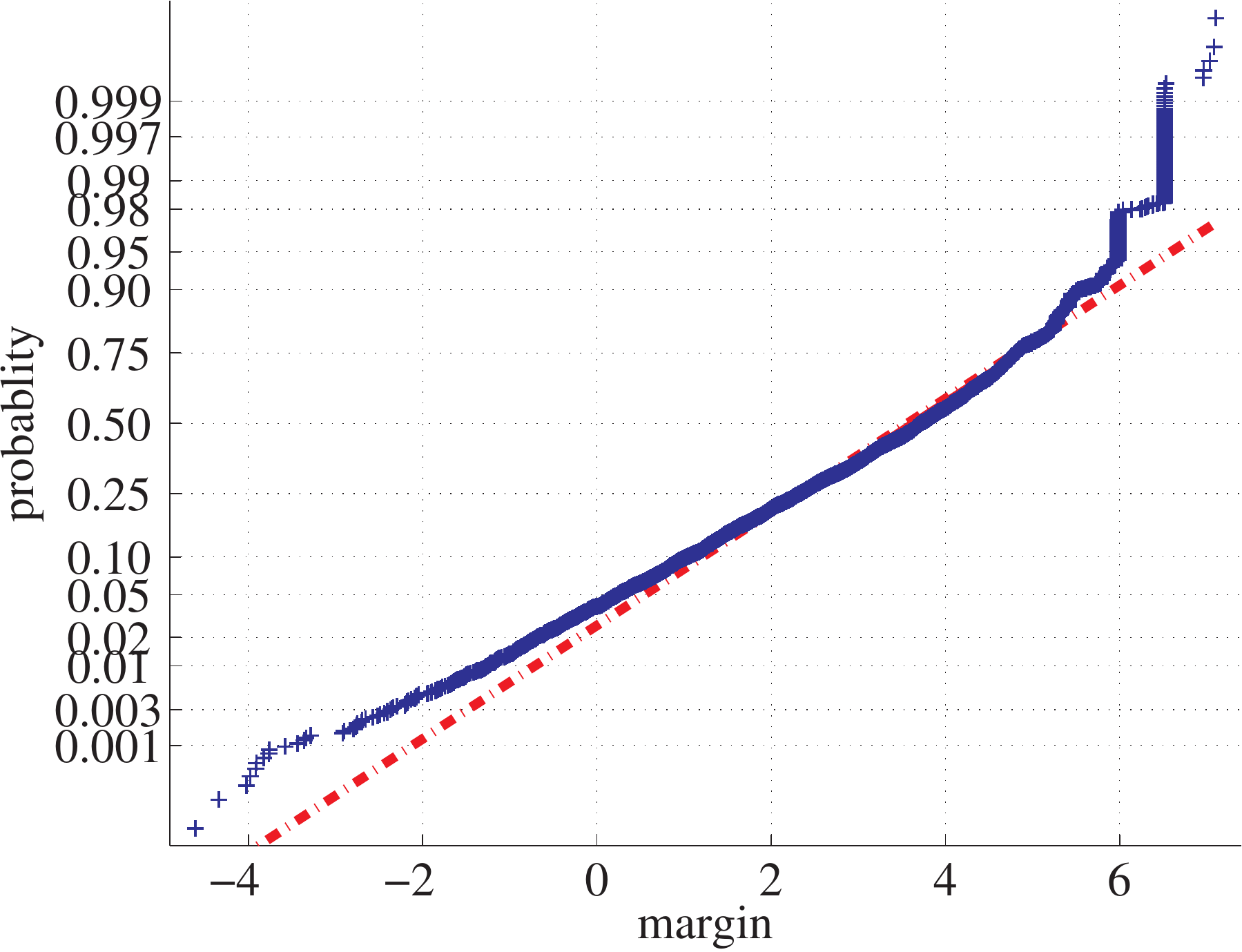}
        \includegraphics[width=.32\textwidth]{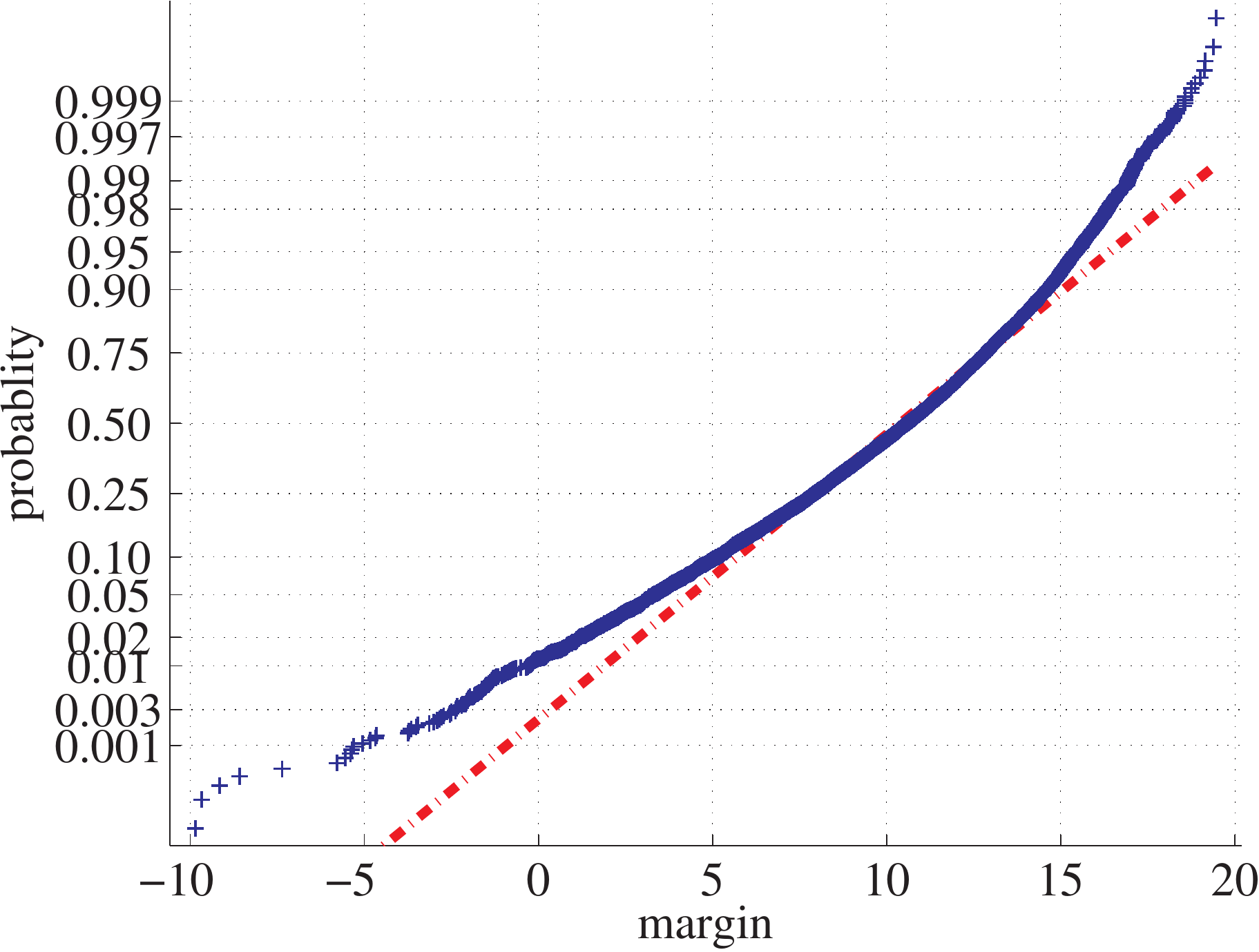}
    \caption{Normality test (normal probability plot)
    for the face data's margin distribution of nodes $1$, $2$, $3$.
    The $ 3 $ nodes contains $ 7 $, $ 22 $, $ 52 $ weak classifiers respectively.
    The data are plotted against a theoretical normal distribution such that
    the data which follows the normal distribution model should form a straight line.
    Curves deviated from the straight line (the red line) indicate
    departures from normality.
    The larger the number of weak classifiers, the more closely the margin follow the
    Gaussian distribution.
    }
    \label{fig:normplot}
\end{figure*}

    In this experiments,
    eight asymmetric boosting methods are evaluated with the multi-exit cascade \cite{pham08multi},
    which are FisherBoost/LACBoost, AdaBoost alone or with LDA/LAC post-processing \cite{wu2008fast},
    AsymBoost alone or with LDA/LAC post-processing.
    We have also implemented Viola-Jones' face detector (AdaBoost with the conventional cascade) 
	as the baseline \cite{viola2004robust}.
    Furthermore, our face detector is also compared with state-of-the-art including some cascade design methods, 
	\ie, WaldBoost \cite{Sochman2005WaldBoost}, FloatBoost \cite{li2004float}, Boosting Chain \cite{xiao2003}
    and the extension of \cite{Saberian2010Boosting}, RCECBoost \cite{Saberian2012Learning}.
    The algorithm for training a multi-exit cascade is summarized in Algorithm \ref{alg:MultiExit_LAC}.

\def\dmin{ d_{\mathrm{min}} }
\def\fmax{ f_{\mathrm{max}} }
\def\ftarget{ F_{\mathrm{fp}} }

\setcounter{AlgoLine}{0}
\linesnumbered\SetVline
\begin{algorithm}[t]
\caption{The procedure for training a multi-exit cascade with LACBoost or FisherBoost.}
\centering   
{\small{
   \begin{minipage}[]{0.94\linewidth}
    \KwIn{
  
        \ADot
         A training set with $m$ examples, which are ordered by their labels 
         ($m_1$ positive examples followed by $m_2$ negative examples);

        \ADot
         $\dmin$:  minimum acceptable detection rate per node;
    
         \ADot
         $\fmax$:  maximum acceptable false positive rate per node;
      
        \ADot
         $ \ftarget $: target overall false positive rate.
         }   
   { {\bf Initialize}:
   \\
   $t = 0$;    ({\em node index}) \\
   $ n = 0 $;  ({\em total selected weak classifiers up to the current node})  
   \\
   $D_t = 1$;  
   $ F_t = 1$. ({\em overall detection rate and false positive 
                     rate up to the current node})
   }

\While{ $  \ftarget < F_t $ }
{
$ t = t + 1$;    ({\em increment node  index})
\\
      \While{
              $ d_t < \dmin  $ 
            }
      { 
       ({\em current detection rate $ d_t $ is not acceptable yet})\\
    \ADot 
        $ n = n + 1$, and generate a weak classifier and update all the weak classifiers'
        linear coefficient 
        using LACBoost or FisherBoost.
       
     \ADot 
       Adjust threshold $b$ of the current boosted strong classifier 
       \[ F^t({\bf x}) =
       \sum_{j = 1}^{n} w^t_j  h_j ({\bf x}) - b\]
       such that $f_t \approx  \fmax$. 
     
     \ADot
        Update the detection rate of the current node 
       $d_t$ with the learned boosted classifier. 
      }
      Update $ D_{t+1} = D_t \times d_t $; $ F_{t+1} = F_t \times f_t $\;

      Remove correctly classified negative samples from negative training set.
      
      \If{ $\ftarget < F_t $ }
      {Evaluate the current cascaded classifier on the negative images
      and add misclassified samples into the negative training set;
      ({\em bootstrap})
      }
}
\KwOut{
      A multi-exit cascade classifier with $n$ weak classifiers
      and $t $ nodes.

}

\end{minipage}
}}
    \label{alg:MultiExit_LAC}
\end{algorithm}

    We first illustrate the validity of adopting LAC and Fisher LDA
    post-processing to improve the node learning objective in the cascade classifier.
    As described abo\-ve, LAC and LDA assume that the margin of the training data
    associated with the node classifier in such a cascade 
    exhibits a Gaussian distribution.
    We demonstrate this assumption on the face detection task in Fig.~\ref{fig:normplot}.
    Fig.~\ref{fig:normplot} shows the normal probability plot 
    of the margins of the positive training
    data for the first three node classifiers in the multi-exit LAC classifier.      %
    The figure reveals that
    the larger the number of weak classifiers used the more closely the margins follow the
    Gaussian distribution.  
    From this, we infer that LAC/LDA post-pro\-ce\-ssing
    and thus LACBoost and FisherBoost, can be
    expected to achieve a better performance when a larger number of weak classifiers are used.
    We therefore apply LAC/LDA only within the later nodes (for example, $9$ onwards)
    of a multi-exit cascade as these nodes contain more weak classifiers.
    We choose multi-exit due to its  property\footnote{
    Since the multi-exit cascade makes use of all previous weak classifiers
    in earlier nodes, it would meet the Gaussianity requirement better than
    the conventional cascade classifier.
    }
    and effectiveness as reported in \cite{pham08multi}.
    We have compared
    the multi-exit cascade with LDA/LAC post-pro\-ce\-ssing against
    the conventional cascade with LDA/LAC post-pro\-ce\-ssing in \cite{wu2008fast}
    and performance improvement has been observed.

    As in \citep{wu2008fast}, five  basic types of Haar-like features
    are calculated, resulting in a $162, 336$ dimensional over-complete
    feature set on an image of $24 \times 24$ pixels.
    To speed up the weak classifier training, as in \cite{wu2008fast},
    we  uniformly sample $10\%$ of features for
    training weak classifiers (decision stumps).
    The face data set consists of $9,832$ mirrored $24 \times 24$ images \cite{viola2004robust}
    ($5,000$ images used for training and $4,832$ imaged used for validation) 
    and $7,323$ larger resolution background images,
    as used in \cite{wu2008fast}. 

    Several multi-exit cascades are trained with various algorithms described above.
    In order to ensure a fair comparison,
    we have used the same number of multi-exit stages and the
    same number of weak classifiers. 
    Each multi-exit cascade consists of $22$ exits and $2,923$ weak classifiers.
    The indices of exit nodes are pre-determined to simplify the training procedure.

    For our FisherBoost and LACBoost, we have an important parameter
    $\theta$, which is chosen from
    $\{
     \frac{1}{10},
     \frac{1}{12},
     \frac{1}{15},
    $
    $
     \frac{1}{20},
     \frac{1}{25},
     \frac{1}{30},
    $
    $
     \frac{1}{40},
     \frac{1}{50}
     \}
    $.
    We have not carefully tuned this parameter using cross-validation.
    Instead, we train a $10$-node cascade for each candidate $ \theta$,
    and choose the one with the best {\em training}
    accuracy.\footnote{To train a complete $22$-node cascade
    and choose the best $ \theta $
    on cross-validation data may give better detection rates.}
    At each exit, negative examples misclassified by current cascade are
    discarded, and new negative examples are bootstrapped from the background
    images pool.
    In total, billions of negative examples are extracted from the pool.
    The positive training data and validation data keep un\-chang\-ed  during the
    training process.

\begin{figure*}[t]
    \begin{center}
        \subfloat[Comparison with asymmetric boosting methods]
          {\includegraphics[width=0.45\textwidth,clip]{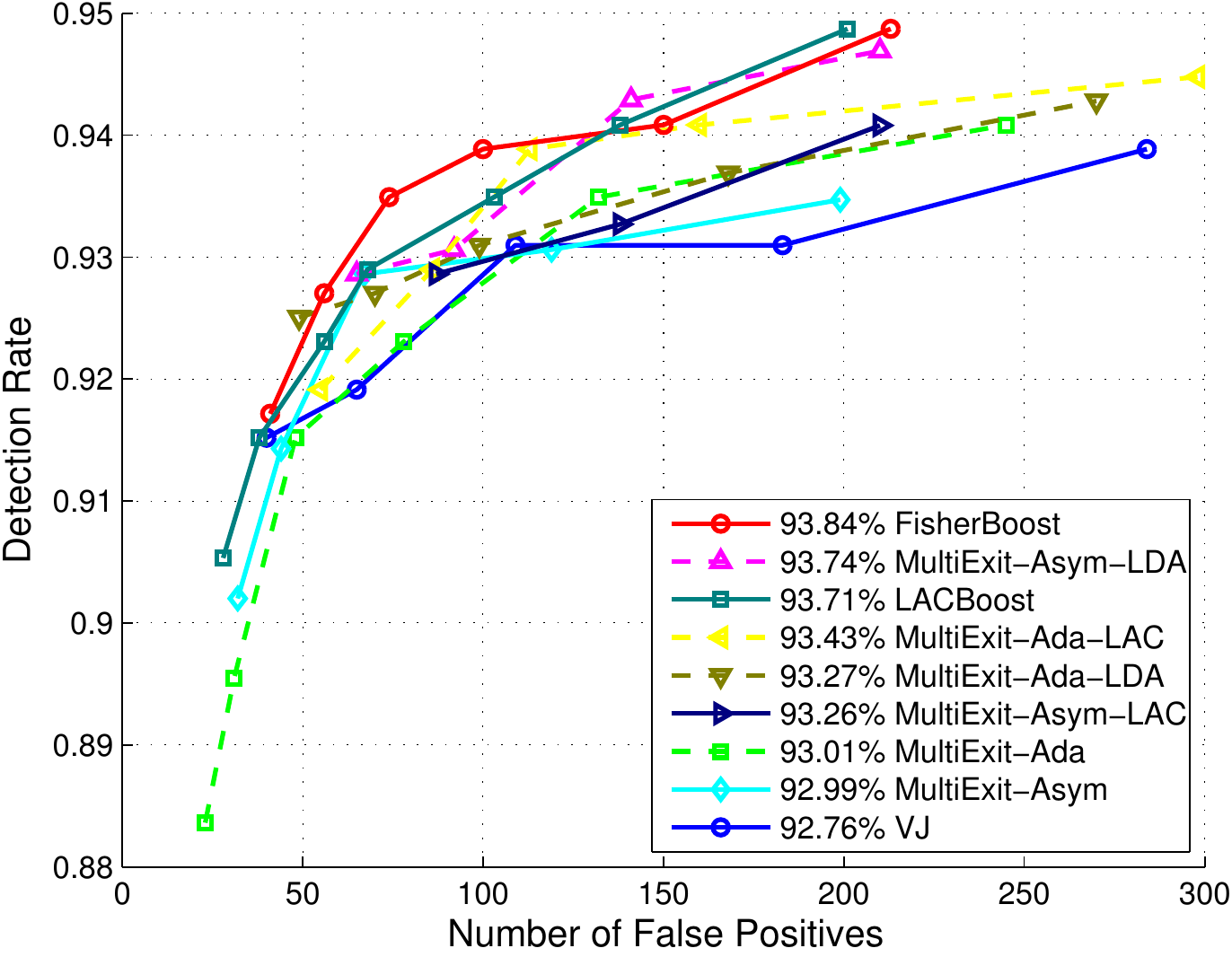}}
        \subfloat[Comparison with some state-of-the-art]
          {\includegraphics[width=0.45\textwidth,clip]{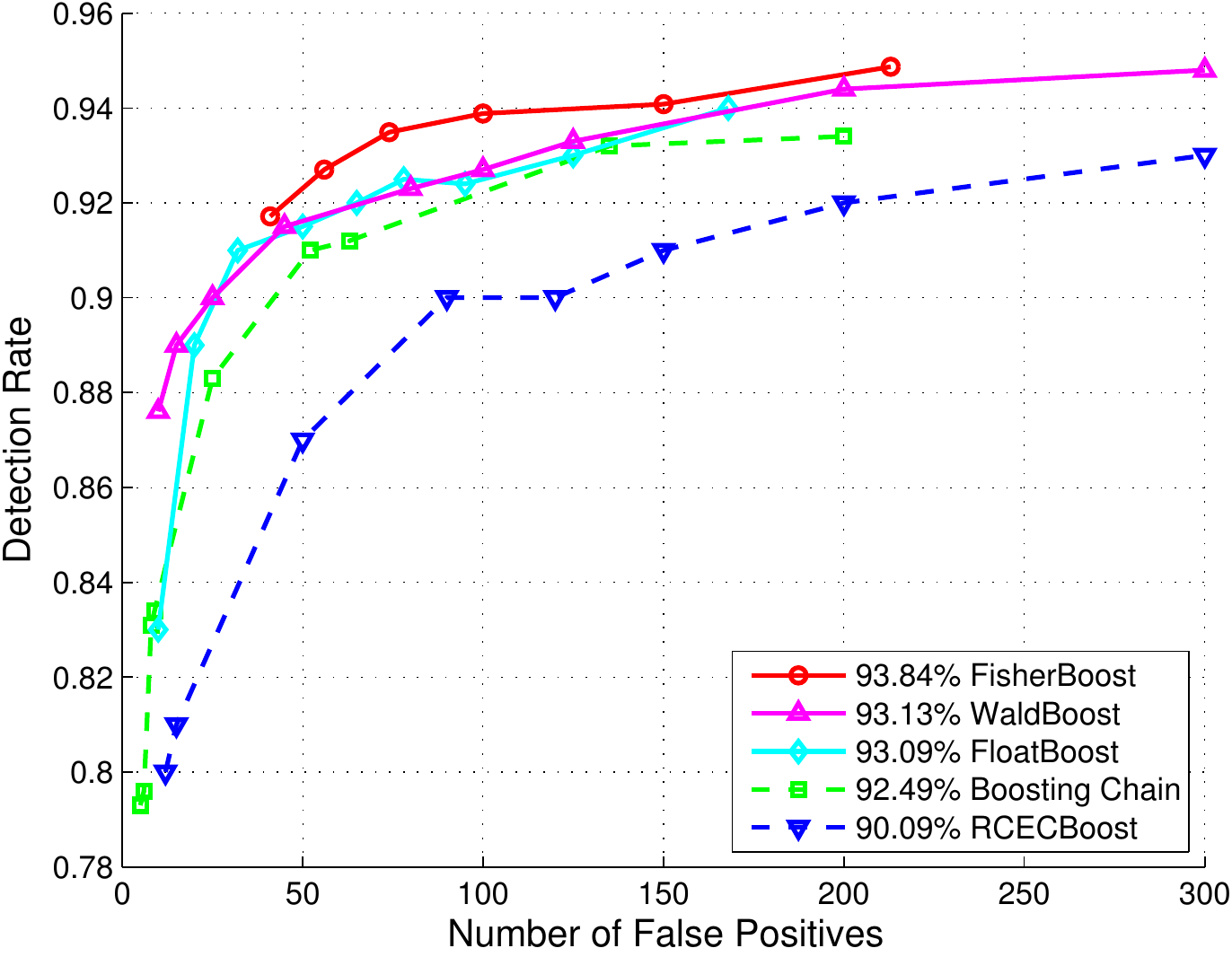}}
    \end{center}
    \caption{
    \modification{
    Our face detectors are compared with other asymmetric boosting methods (a) and some state-of-the-art including cascade design methods (b)
    on MIT+CMU frontal face test data using ROC curves (number of false positives versus detection rate).
    ``Ada'' and ``Asym'' mean that features are selected using AdaBoost and AsymBoost, respectively.
    ``VJ'' implements Viola and Jones' cascade using AdaBoost \cite{viola2004robust}.
    ``MultiExit'' means the multi-exit cascade \cite{pham08multi}.
    The ROC curves of compared methods in (b) are quoted
    from their original papers \cite{Sochman2005WaldBoost,li2004float,xiao2003,Saberian2012Learning}.
	Compared methods are ranked in the legend, based on the average of detection rates.
    }}
    \label{fig:ROC1}
\end{figure*}

    Our experiments are performed on a workstation with $8$ Intel Xeon
    E$5520$ CPUs and $32$GB RAM.
    It takes about $3$ hours  to train the multi-exit cascade with AdaBoost or AsymBoost.
    For FisherBoost and LACBoost, it takes less than $ 4 $ hours to train
    a complete multi-exit cascade.\footnote{Our implementation is in C++ and only the weak classifier
              training part is parallelized using OpenMP. 
              }
    In other words,
    our EG algorithm takes less than $ 1 $ hour to
    solve the primal QP problem (we need to solve a QP at each iteration).
    As an estimation of the computational complexity,
    suppose that the number of training
    examples is $  m $, number of weak classifiers is $ n $.
    At each iteration of the cascade training,
    the complexity of solving the primal QP using EG is
    $  O( m n  + k n^2) $ with $ k $ the iterations
    needed for EG's convergence.
    The complexity  for  training the weak classifier  is
    $ O( m  d ) $ with $d$ the number of all Haar-feature patterns.
    In our experiment, $ m = 10,000 $,
    $ n \approx 2900 $,
    $d = 160,000$,
    $ k < 500 $.
    So the majority of the computational cost of the training process
    is bound up in the weak classifier training.

    We have also experimentally observed the speedup of EG against standard QP solvers.
    We solve the primal QP defined by \eqref{EQ:QP2} using EG and Mosek \cite{Mosek}.
    The QP's size is $ 1,000 $ variables.
    With the same accuracy tolerance (Mosek's primal-dual gap is set to $ 10^{-7}$
    and EG's convergence tolerance is also set to $ 10^{-7}$),
    Mosek takes $1.22 $ seconds and EG is
    $ 0.0541 $ seconds on a standard desktop.
    So EG is about $ 20 $ times faster.
    Moreover,  at iteration $ n + 1 $ of training the cascade,
    EG can take advantage of the last iteration's solution
    by starting EG from a small perturbation of the previous solution.
    Such a warm-start gains a $ 5 $ to $ 10\times $ speedup in our experiment,
    while the current QP solver in Mosek does not support
    warm-start  \citep[Chapter 7]{Mosek}.

    We evaluate the detection performance on the MIT\-+\-CMU frontal
    face test set.
    This dataset is made up of $507$ frontal faces in $130$ images with
    different background.

    If one positive output has less than $50\%$ variation of shift and
    scale from the ground-truth, we treat it as a true positive, otherwise
    a false positive.

    In the test phase,
    the scale factor of the scanning window is set to $1.2$ and
    the stride step is set to $1$ pixel.

    The Receiver operating characteristic (ROC) curves in Fig.~\ref{fig:ROC1} 
	show the entire cascade's performance.
	The average detection rate (similar with the one used in \cite{Dollar2012Pedestrian}) 
	are used to rank the compared methods,
	which is the mean of detection rates sampled evenly from $50$ to $200$ false positives.	
    Note that multiple factors impact on the cascade's performance, however,
    including: the classifier set,
    the cascade structure,  bootstrapping {\em etc}.
    Fig.~\ref{fig:ROC1} (a) demonstrate the superior performance of FisherBoost
    to other asymmetric boosting methods in the face detection task.
    We can also find that LACBoost perform worse than FisherBoost.
    Wu {et al.} have observed that LAC post-processing
    does not outperform LDA post-processing in some cases either.
    We have also compared our methods with the boost\-ed greedy sparse LDA (BGSLDA) in
    \cite{paul2009cvpr,GSLDA2010Shen},
    which is considered one of the state-of-the-art.
    FisherBoost and LACBoost outperform
    BGSLDA with Ada\-Boost\-/Asym\-Boost 
    in the detection rate.
    Note that BGSLDA uses the standard cascade.

    From Fig.~\ref{fig:ROC1} (b), we can see the performance of FisherBoost is
    better than the other considered cascade design methods.
    However, since the parameters of cascade structure (\eg, node thresholds, number of nodes, number of weak classifiers per node)
    are not carefully tuned,
    our method can not guarantee an optimal trade-off between accuracy and speed.
    We believe that the boosting method and the cascade design
    strategy compensate each other.  
    Actually in \cite{Saberian2010Boosting}, the authors also incorporate some cost-sensitive boosting algorithms,
    \eg, cost-sensitive AdaBoost \cite{Masnadi2011Cost}, AsymBoost \cite{Viola2002Fast},
    with their cascade design method.

\subsection{Pedestrian Detection Using a Cascade Classifier}
\label{ped_cascade}

    We run our experiments on a pedestrian detection with a minor
    modification to visual features being used.
    We evaluate our approach on INRIA data set \cite{Dalal2005HOG}.
    The training set consists of $2,416$ cropped mirrored pedestrian images and $1,200$ large
    resolution background images.
    The test set consists of $288$ images containing $588$ annotated pedestrians and $453$
    non-pedestrian images.
    Each training sample is scal\-ed to $64 \times 128$ pixels with an additional of $16$
    pixels added to each border to preserve human contour information.
    During testing, the detection scanning window is resized to $32 \times 96$ pixels
    to fit the human body.
    We use histogram of oriented gradient (HOG) features in our experiments.
    Instead of using fixed-size blocks ($105$ blocks of size $16 \times 16$ pixels) as in
    Dalal and Triggs \cite{Dalal2005HOG},
    we define blocks with various scales (from $12 \times$ 12 pixels
    to $64 \times 128$ pixels)
    and width-length ratios ($1:1$, $1:2$, $2:1$, $1:3$, and $3:1$).
    Each block is divided into $2 \times 2$ cells, and HOG features
    in each cell are summarized into $9$ bins.
    Hence $36$-dimensional HOG feature is generated from each block.
    In total, there are $ 7,735 $ blocks from a $64 \times 128$-pixels patch.
    $ \ell_1$-norm normalization is then applied to the feature vector.
    Furthermore, we use integral histograms to speed up the computation as in \cite{zhu2006fast}.
    At each iteration, we randomly sample $10\%$ of all the possible blocks for training a weak
    classifier.
    We have used weighted linear discriminant analysis (WLDA) as weak classifiers, same as in
    \cite{paul2008fast}. Zhu et al. used linear support vector machines as weak
    classifiers \cite{zhu2006fast}, which can also be used as weak classifiers here.

\begin{figure}[t]
  \begin{center}
    \includegraphics[width=0.45\textwidth,clip]{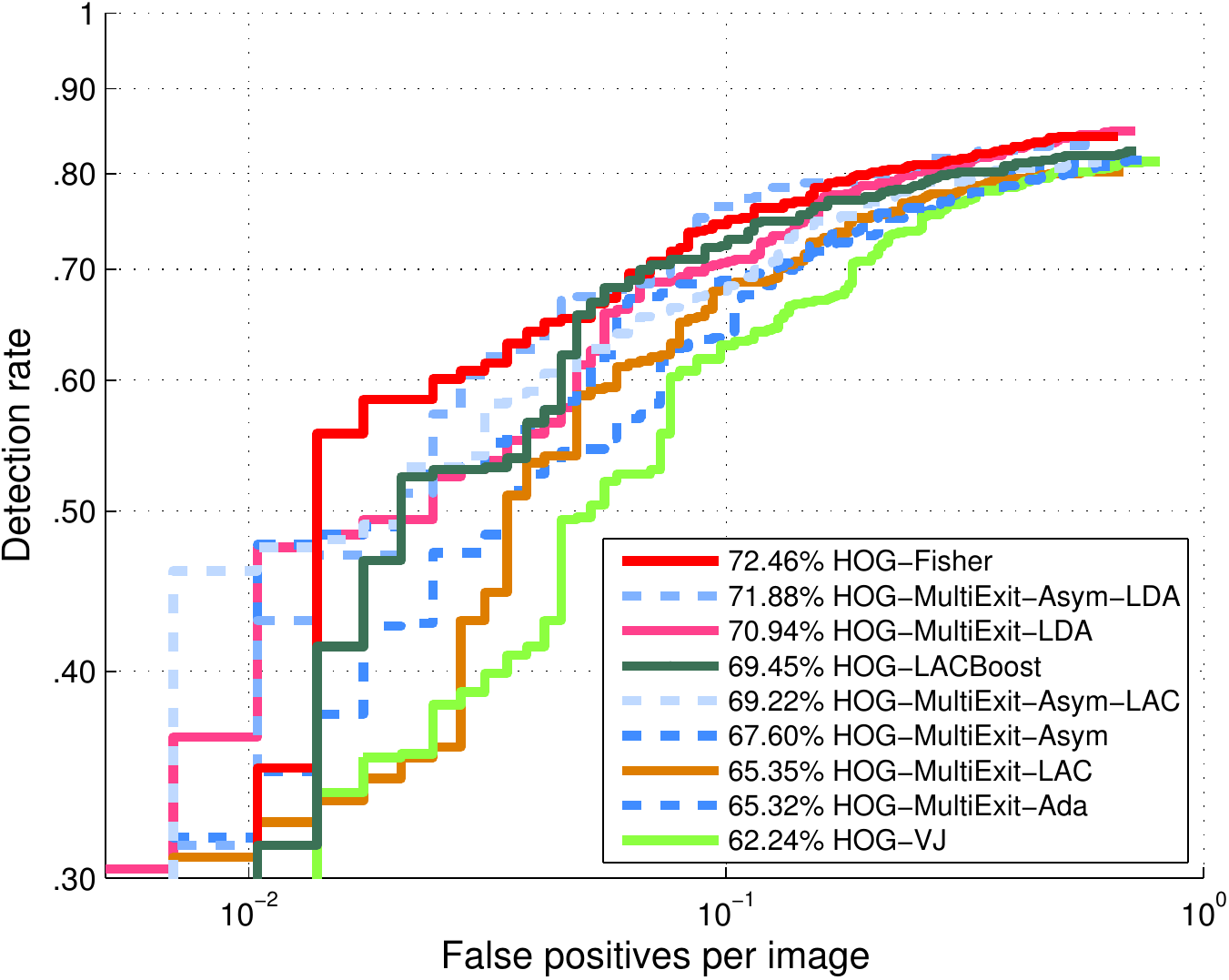}
  \end{center}
  \caption{
  \modification{
  FisherBoost (HOG-Fisher) and LACBoost (HOG-LACBoost) are compared
  with other cascade pedestrian detectors on the INRIA data set.
  All cascades are trained with the same number of weak classifiers and nodes, using HOG features.
  In the legend, detectors are sorted based on their log-average detection rates. FisherBoost performs best compared to other cascades.}}
  \label{fig:responsesC0}
\end{figure}

    In this experiment, all cascade classifiers have  	
	the same number of nodes and weak classifiers.
    For the same reason described in the face detection section,
    the FisherBoost/LACBoost and Wu et al.'s LDA/LAC post-pro\-ce\-ssing
    are applied to the cascade from the $3$-rd node onwards, instead of the first node.
    The positive examples remain the same for all nodes
    while the negative examples in later nodes are obtained by a bootstrap approach.
    The parameter $\theta$ of our FisherBoost and LACBoost is selected from
     $\{
      \frac{1}{10}$,
      $\frac{1}{12}$,
      $\frac{1}{14}$,
      $\frac{1}{16}$,
      $\frac{1}{18}$,
      $\frac{1}{20}
     \}$.
    We have not carefully selected $ \theta $ in this experiment.
    Ideally, cross-validation should be used to pick the best value of $ \theta $ by using
    an independent cross-validation data set.
    Since there are not many labeled positive training data in the INRIA data set,
    we use the same $2,416$ positive examples for validation.
    We collect $500$ additional negative examples by bootstrapping for validation.
    Further improvement is expected if the positive data used during validation is different from
    those used during training.
    During evaluation, we use a step stride of $4 \times 4$ pixels with $10$ scales per octave (a scale ratio of $1.0718$).
    The performance of different cascade detectors is evaluated using a protocol described in \cite{Dollar2012Pedestrian}.
    A technique known as pairwise maximum suppression \cite{PMT}
    is applied to suppress less confident detection windows.
	A confidence score is needed for each detection window as the input of pairwise maximum suppression.
	In this work, this confidence is simply calculated as the mean of decision scores of the last five nodes in the cascade.

    The ROC curves are plotted in Fig.~\ref{fig:responsesC0}.
	Same as \cite{Dollar2012Pedestrian}, the log-average detection rate is used to summarize overall detection performance, 
	which is the mean of detection rates sampled evenly at $9$ positions from $0.01$ to $1$.	
    In general, FisherBoost (HOG-Fisher) outperforms all other cascade detectors.
    Similar to our previous experiments, LAC and LDA post-processing
    further improve the performance of AdaBoost.
    However, we observe that both FisherBoost and LDA post-processing
    have a better generalization performance than LACBoost and LAC post-processing.
	We will discuss this issue at the end of the experiments.

{
\subsection{Comparison with State-of-the-art Pedestrian Detectors}

In this experiment, we compare FisherBoost with state-of-the-art pedestrian detectors on several public data sets.
    In \cite{Dollar2012Pedestrian}, the authors compare various
    pedestrian detectors and conclude that combining multiple
    discriminative features can  often significantly boost the
    performance of pedestrian detection.  This is not surprising since
    a similar conclusion was drawn in \cite{Gehler2009Feature}  on an
    object recognition task.  Clearly, the pedestrian detector, which
    relies solely on the HOG feature, is unlikely to outperform those
    using a combination of features.

        To this end, we train our pedestrian detector by combining
        both HOG features \cite{Dalal2005HOG} and covariance features
        \cite{Tuzel2008PAMI}\footnote{ Covariance features capture the
        relationship between different image statistics and have been
        shown to perform well in our previous experiments.  However,
        other discriminative features can also be used here instead,
        \eg, Haar-like features, Local Binary Pattern (LBP)
        \cite{mu2008lbp} and self-similarity of low-level features
        (CSS) \cite{Walk2010New}.}.
        For HOG, we use the same experimental settings as our previous
        experiment.  For covariance features, we use the following
        image statistics
        $ \Big[ x,$ 
            $y,$ 
            $I,$ $| I_x |,$ $| I_y |,$
            ${\textstyle \sqrt{I_x^2 + I_y^2}},$ $| I_{xx} |, $
            $| I_{yy} |,$ 
            $\textrm{arctan} (| I_x | / | I_y |) \Big] $,
where $x$ and $y$ are the pixel location,
$I$ is the pixel intensity,
$I_x$ and $I_y$ are first order intensity derivatives,
$I_{xx}$ and $I_{yy}$ are second order intensity derivatives and
the edge orientation.
    Each pixel is mapped to a $9$-dimensional feature image.  We then
    calculate $36$ correlation coefficients in each block and
    concatenate these features to previously computed HOG features.
    The new feature not only encodes the gradient histogram (edges)
    but also information of the correlation of defined statistics
    inside each spatial layout (texture).  Similar to the previous
    experiment, we project these new features to a line using weighted
    linear discriminant analysis.  Except for new features, other
    training and test implementations are the same with those in the
    previous pedestrian detection experiments.

\begin{figure*}[t]
  \centering
    \subfloat[INRIA]{
        \includegraphics[width=0.45\textwidth,clip]{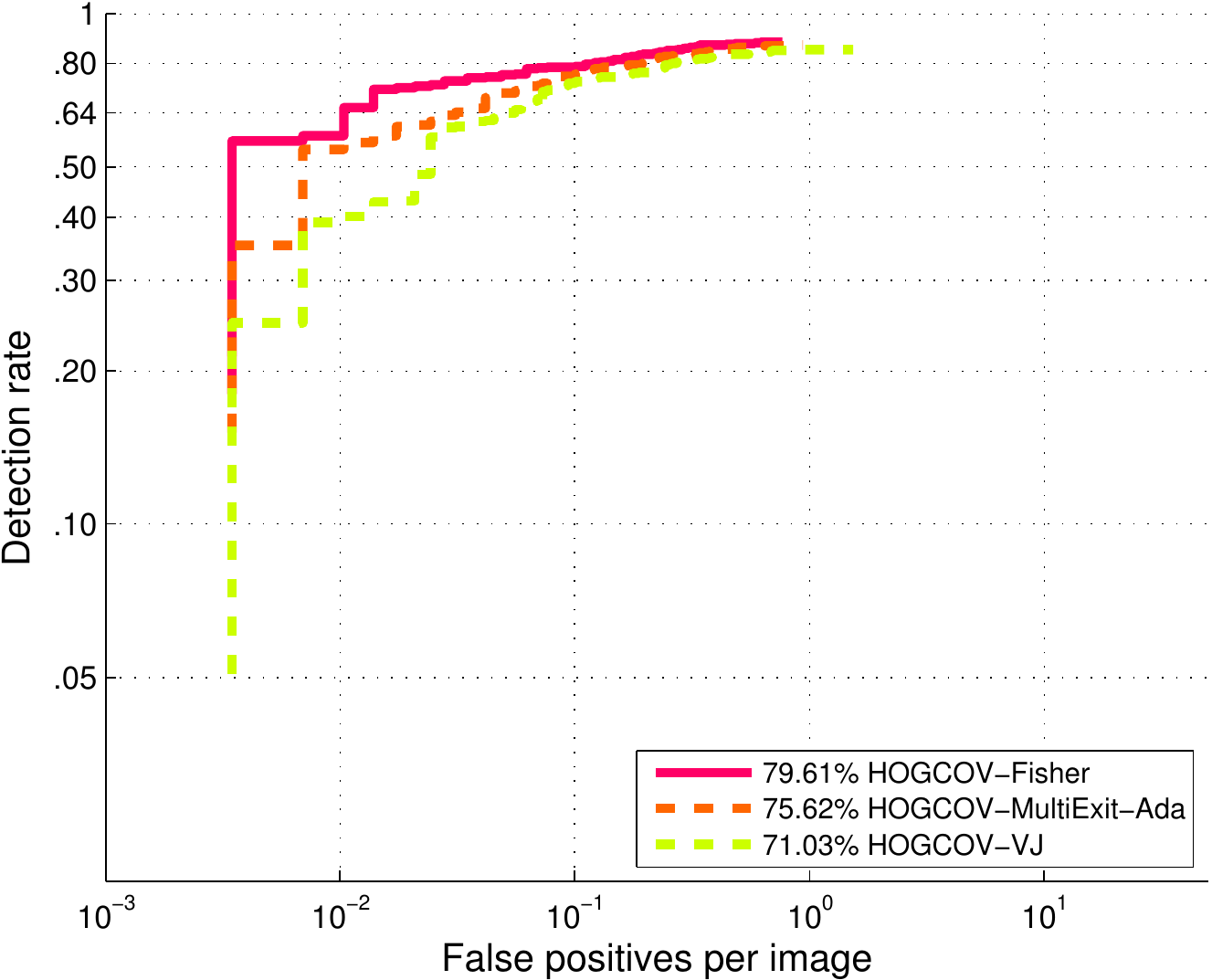}
        \label{fig:responseB1A}
    }
    \qquad
     \subfloat[INRIA]{
        \includegraphics[width=0.45\textwidth,clip]{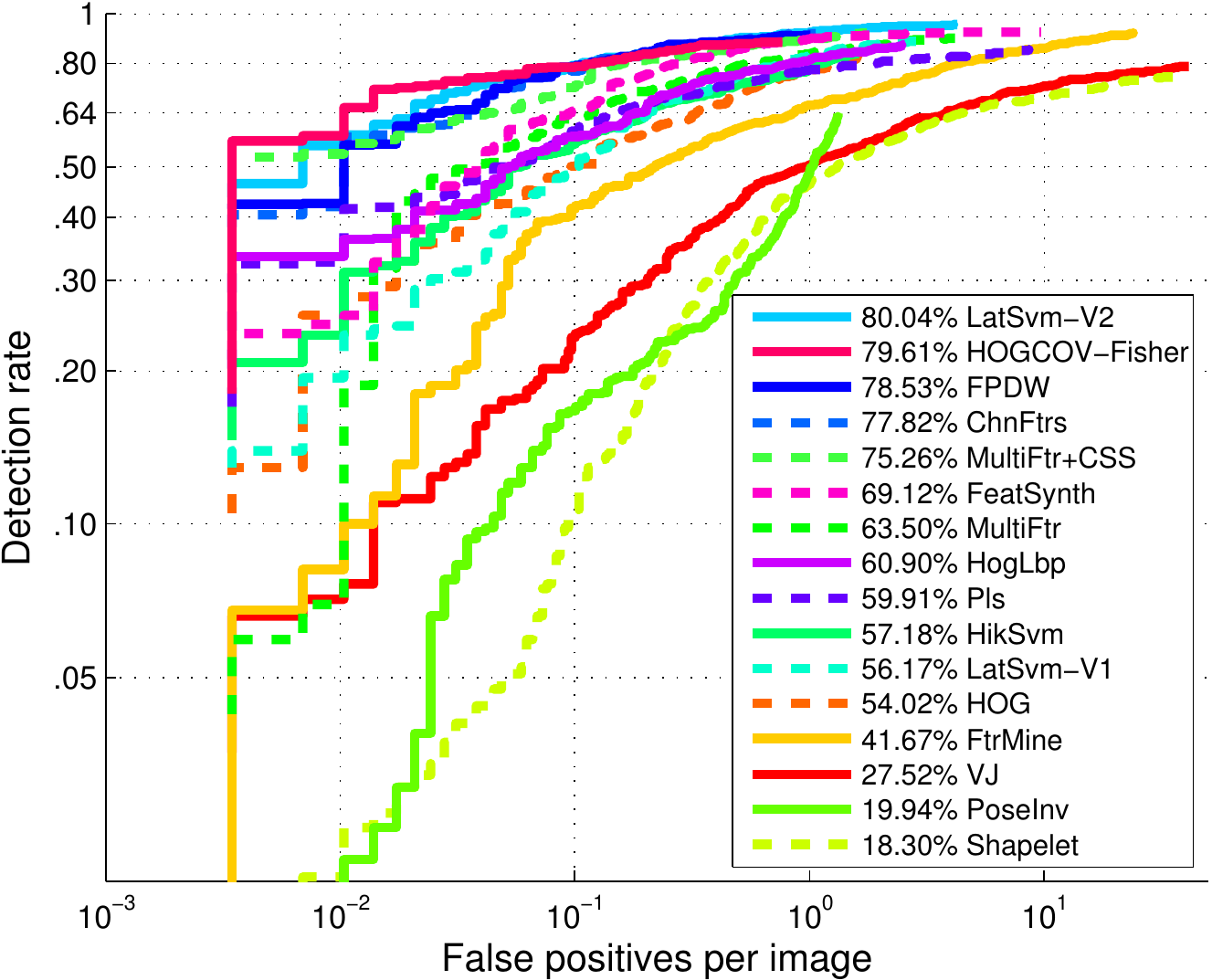}
        \label{fig:responseB1B}
    }
    \qquad
    \subfloat[TUD-Brussels]{
        \includegraphics[width=0.45\textwidth,clip]{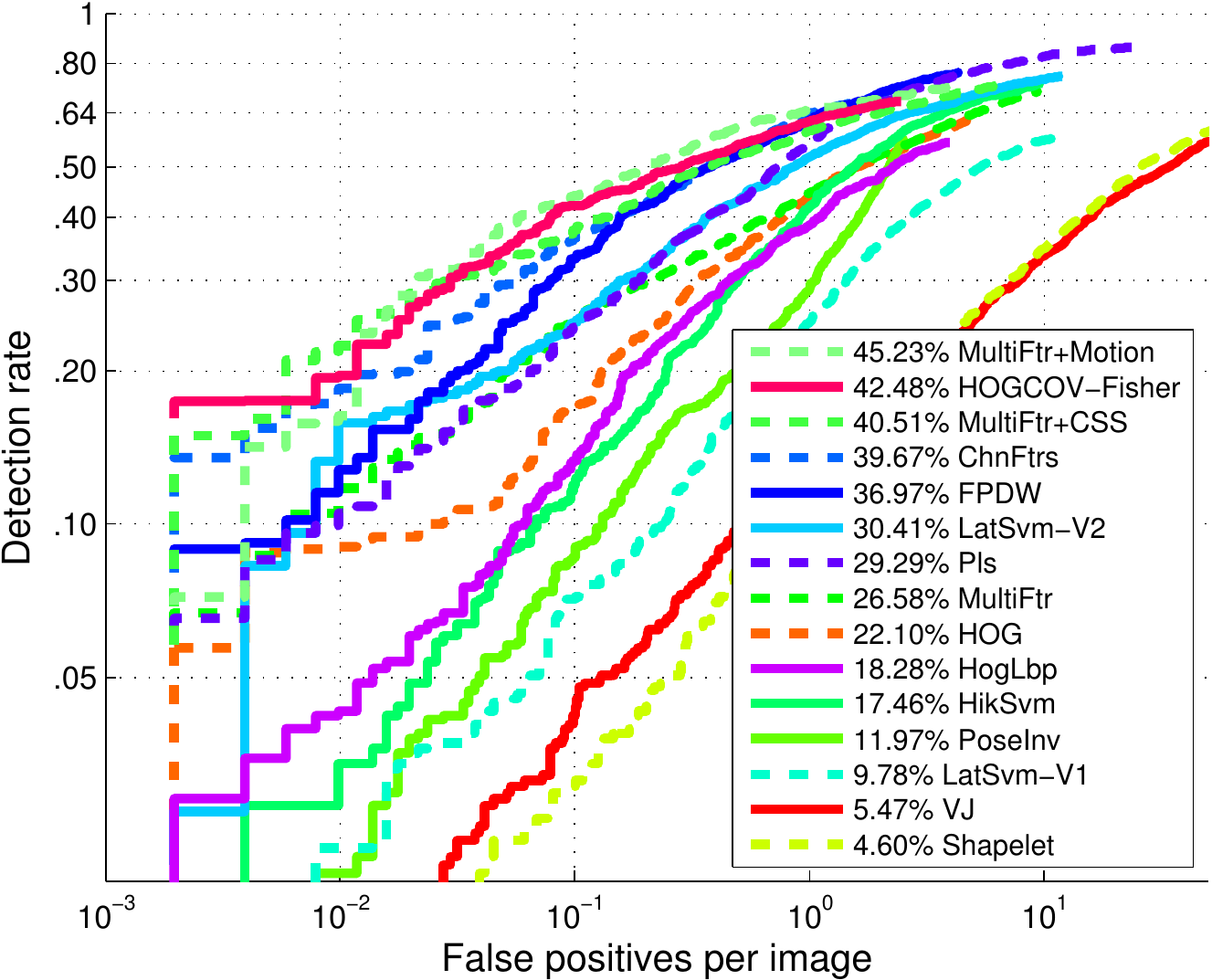}
        \label{fig:responseB1C}
    }
    \qquad
    \subfloat[ETH]{
        \includegraphics[width=0.45\textwidth,clip]{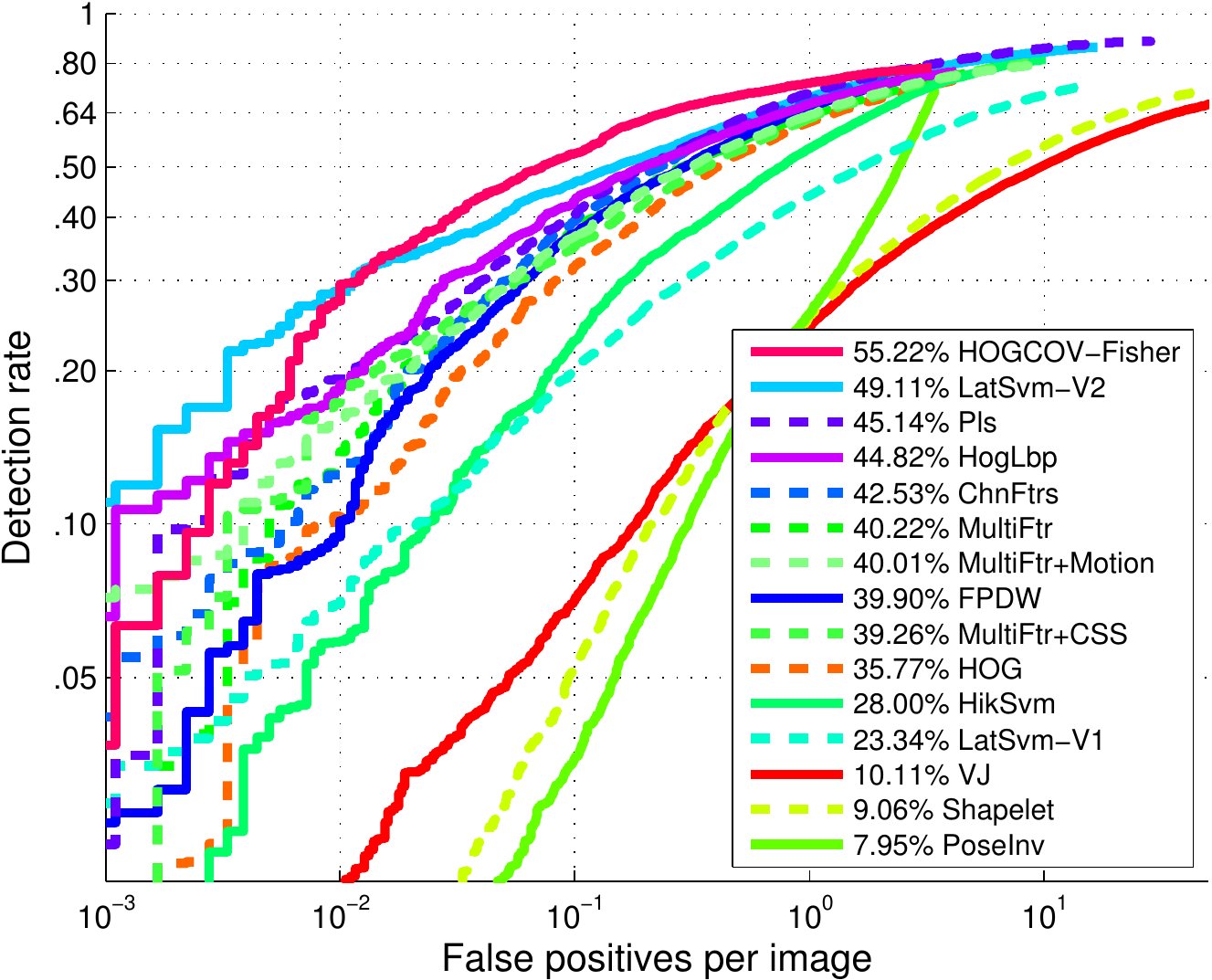}
        \label{fig:responseB1D}
    }
  \caption{
        The performance of our pedestrian detector (HOGCOV-Fisher)
        compared with (a) baseline detectors and (b, c, d)
        state-of-the-art detectors on publicly available pedestrian
        data sets. Our detector uses HOG and covariance features.  The
        performances are ranked using log-average detection rates in
        the legend.  Our detector performs best on the INRIA
        \cite{Dalal2005HOG} data set, second best on the
        TUD-Bru\-ss\-els \cite{Wojek2009} and ETH \cite{ess2007} data
        sets. Note that the best one on the latter two data sets has
        either used many more features or used a more sophisticated
        part-based model.    
  }
  \label{fig:responsesB1}
\end{figure*}

We first compare FisherBoost (HOGCOV-Fisher) with two baseline detectors trained with AdaBoost.
The first baseline detector is trained with the conventional cascade (HOGCOV-VJ) 
while the second baseline detector is trained with the multi-exit cascade (HOGCOV-MultiExit-Ada).
All detectors are trained with both HOG and covariance features on INRIA training set.
The results on INRIA test sets using the protocol in \cite{Dollar2012Pedestrian} are reported in Fig.~\ref{fig:responsesB1} (a).
Similar to previous results, FisherBoost outperforms both baseline detectors.

Our detector is then compared with existing pedestrian detectors listed in \cite{Dollar2012Pedestrian},
on the INRIA, TUD-Brussels and ETH data sets.
For the TUD-Bru\-ss\-els and ETH data sets, since sizes of ground-truths are smaller than that in INRIA training set, 
we upsample the original image to $1280 \times 960$ pixels before applying our pedestrian detector.
ROC curves and log-average detection rates are reported in Fig.~\ref{fig:responsesB1} (b), (c) and (d).
    On the ETH data set, FisherBoost outperforms all the other $14$
    compared detectors.  On the TUD-B\-ru\-s\-s\-el\-s data set, our detector is the second
    best, only inferior to MultiFtr+\-Motion \cite{Walk2010New} that
    uses more discriminative features (gradient, self-similarity and
    motion) than ours.  On the INRIA data set, FisherBoost's performance is
    also ranked the second, and only worse than the part-based
    detector  
    \cite{Felzenszwalb2008Object} which uses a much more complex model
    (deformable part models) and training process (latent S\-V\-M).  We
    believe that by further combining with more discriminative
    features, \eg, CSS features as used in \cite{Walk2010New}, the
    overall detection performance of our method can be further improved.
    In summary, 
    despite the use of simple HOG plus covariance features,  
    our FisherBoost pedestrian detector still achieves the
    state-of-the-art performance on public benchmark data sets. 

    Finally, we report an average number of features evaluated per
    scanning window in Table~\ref{tab:responsesB2}.  We compare
    FisherBoost with our implementation of AdaBoost with the
    traditional cascade and AdaBoost with the multi-exit cascade.
    Each image is scanned with $4 \times 4$ pixels step stride and
    $10$ scales per octave.  There are $90,650$ patches to be
    classified per image.  On a single-core Intel i$7$ CPU $2.8$ GHz
    processor, our detector achieves an average speed of $0.186$
    frames per second (on $640 \times 480$ pixels CalTech images),
    which is ranked eighth compared with $15$ detectors evaluated in
    \cite{Dollar2012Pedestrian}.  Currently, $90\%$ of the total
    evaluation time is spent on extracting both HOG and covariance
    features $(60\%$ of the evaluation time is spent on extracting raw
    HOG and covariance features while another $30\%$ of the evaluation
    time is spent on computing integral images for fast feature
    calculation during scanning phase).

The major bottleneck of our pedestrian detector lies in the feature extraction part.
In our implementation, we make use of multi-threading to speed up the runtime of our pedestrian detector.
Using all $8$ cores of Intel i$7$ CPU, we are able to speed up an average processing time to less than $1$ second per frame.
We believe that by using a special purpose hardware, such as Graphic Processing Unit (GPU), the speed of our detector can be significantly improved.
}

\begin{table}
  \centering
  \begin{tabular}{r|ccc}
  \hline
         &  avg. features  & frames/sec. \\
  \hline
  \hline
  FisherBoost + multi-exit & $10.89$  & $0.186$ \\
  AdaBoost + multi-exit & $11.35$ & $0.166$ \\
  AdaBoost + VJ cascade & $21.00$  & $0.109$ \\
  \hline
  \end{tabular}
  \caption{
  \modification{
  Average features required per detection window and
  average frames processed per second for different pedestrian 
  detectors on  CalTech images of $640 \times 480$ pixels
  (based on our own implementation).
  }}
  \label{tab:responsesB2}
\end{table}

\modification{

\subsubsection{Discussion}

\paragraph{Impact of varying the number of weak classifiers}
In the next experiment, we vary the number of weak classifiers in each cascade
node to evaluate their impact on the final detection performance.
We train three different pedestrian detectors (Fisher$4$/$5$/$6$, see
Table \ref{tab:responsesB0} for details) on the INRIA data set.
We limit the maximum number of weak classifiers in each multi-exit node to be $80$.
The first two nodes is trained using AdaBoost and subsequent nodes are trained using FisherBoost.
Fig.~\ref{fig:responsesB0} shows ROC curves of different detectors.
Although we observe a performance improvement as the number of weak classifiers increases,
this improvement is minor compared
to a significant increase in the average number of features 
required per detection window.
This experiment indicates the robustness of FisherBoost to the number of weak classifiers in the multi-exit cascade.
Note that Fisher$5$ is used in our previous experiments on pedestrian detection.

\paragraph{Impact of training FisherBoost from an early node}
In the previous section, we conjecture that FisherBoost performs well when the margin follows the Gaussian distribution.
As a result, we apply FisherBoost in the later node of a multi-exit cascade (as these nodes often contain a large number of weak classifiers).
In this experiment, we show that it is possible to start training FisherBoost from the first node of the cascade.
To achieve this, one can train an additional $50$ weak classifiers in the first node (to guarantee the margin approximately follow the Gaussian distribution).
We conduct an experiment by training two FisherBoost detectors.
In the first detector (Fisher$50$), FisherBoost is applied from the first node onwards.
The number of weak classifiers in each node is $55$, $60$ (with $55$
weak classifiers from the first node), $70$ ($60$ weak classifiers
from previous nodes), $80$ ($70$ weak classifiers from previous
nodes), {\em etc}.
In the second detector (Fisher$5$), we apply AdaBoost in the first two
nodes and apply FisherBoost from the third node onwards.  The number
of weak classifiers in each node is $5$, $10$ (with $5$ weak
classifiers from the first node), $20$ ($10$ from previous nodes),
$30$ ($20$ from previous nodes), {\em etc}.
Both detectors use the same
node criterion, \ie, each node should discard at least $50\%$
background samples.  All other configurations are kept to be the same.

\begin{figure*}[t]
  \begin{center}
      \subfloat[]{
    \includegraphics[width=0.45\textwidth,clip]{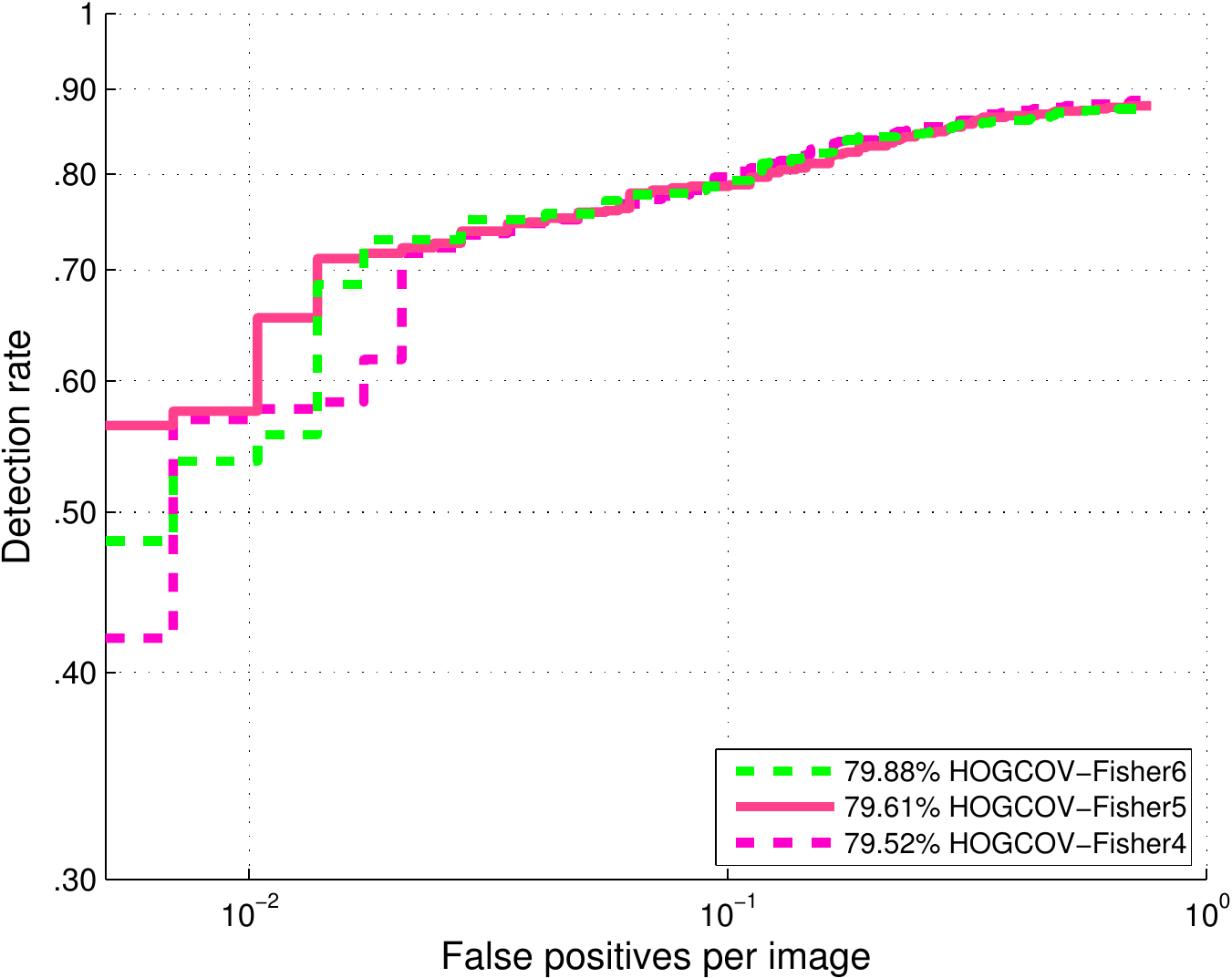}
    }
    \subfloat[]{
    \includegraphics[width=0.45\textwidth,clip]{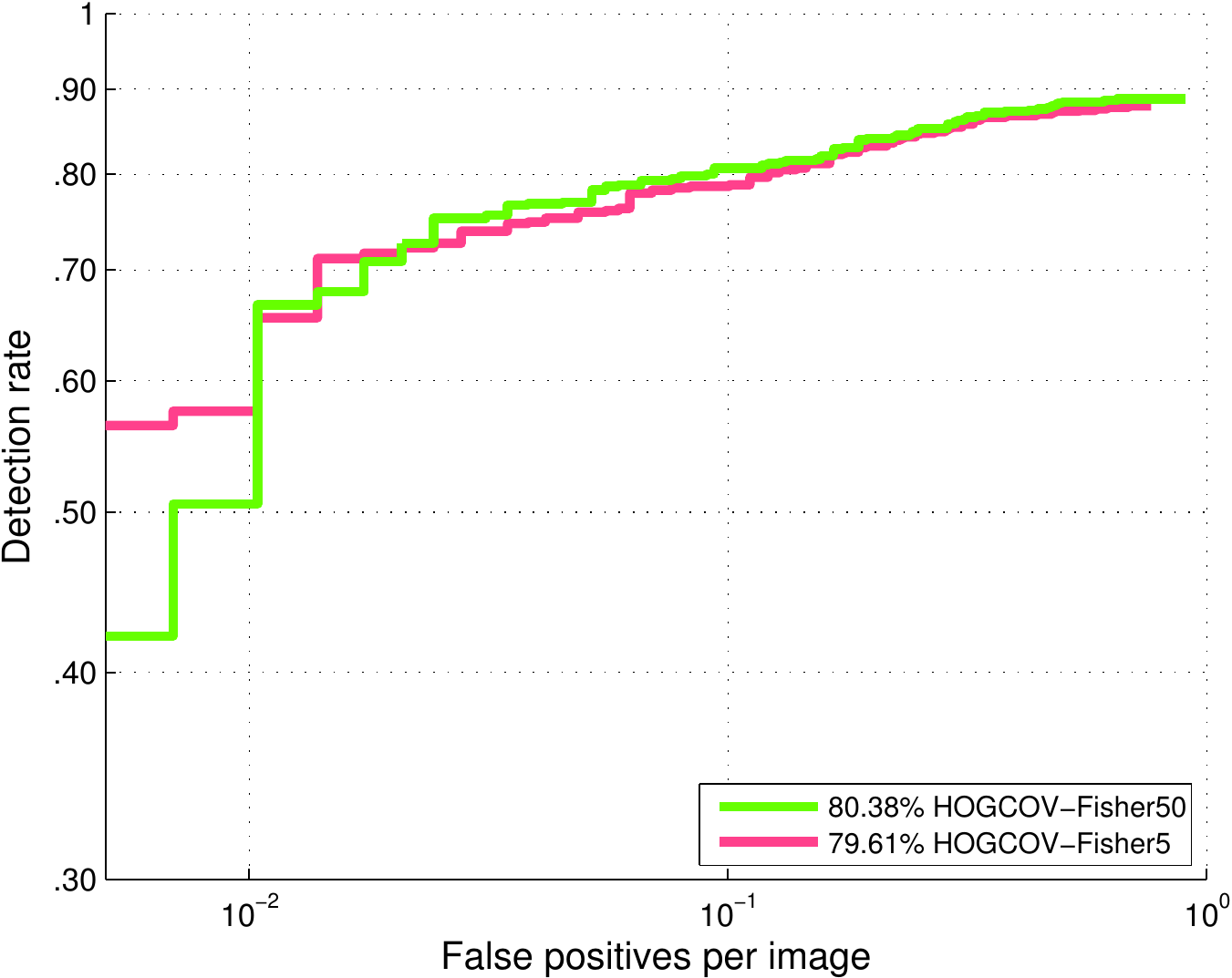}
    }
  \end{center}
 \caption{
  Performance comparison.
  (a) We vary the number of weak classifiers in each multi-exit node.
  When more weak classifiers are used in each node, 
  the accuracy can be slightly improved.
  (b) We start training FisherBoost from the first node (HOGCOV-Fisher$50$).
  HOGCOV-Fisher$50$ can achieve a slightly better detection rate
  than HOGCOV-Fisher$5$.  
  }
  \label{fig:responsesB0}
\end{figure*}

\begin{table*}[t]
  \begin{center}
  \begin{tabular}{l|cccccccccccc|c|c}
  \hline
      & Node $1$ &  $2$ & $3$ & $4$ & $5$ & $6$ & $7$ & $8$ & $9$ &
      $10$ & $11$ & $12$ onwards & avg. features & log-average det.
      rate\\
  \hline
  \hline
    Fisher$4$ &  $4$ & $4$  & $8$  & $8$  & $16$ & $16$ & $32$ & $32$ & $64$ & $64$ & $80$ & $80$   & $26.4$  & $79.52\%$ \\
    Fisher$5$ &  $5$ &  $5$ & $10$ & $10$ & $20$ & $20$ & $40$ & $40$ & $80$ & $80$ & $80$ & $80$   & $26.2$  & $79.61\%$ \\
    Fisher$6$ &  $6$ &  $6$ & $12$ & $12$ & $24$ & $24$ & $48$ & $48$ & $80$ & $80$ & $80$ & $80$   & $30.6$  & $79.88\%$ \\
  \hline
  \end{tabular}
  \end{center}
  \caption{
  We compare the performance of FisherBoost by varying the number of weak classifiers in each multi-exit node.
  Average features required per detection window and log-average
  detection rates on the INRIA pedestrian dataset are reported.
  When more weak classifiers in each multi-exit node are used,
  slightly improved accuracy can be achieved at the price of more
  features being evaluated.
  }
  \label{tab:responsesB0}
\end{table*}

We report the performance of both detectors in Fig. \ref{fig:responsesB0}.
From the results, Fisher$50$ performs slightly better than Fisher$5$ 
(log-average detection rate of $80.38\%$ vs. $79.61\%$).
    Based on these results, classifiers in early nodes of the cascade
    may be heuristically chosen such that a large number of easy
    negative patches can be quickly discarded.  In other words, the
    first few nodes can significantly affect the efficiency of the
    visual detector but do not play a significant role in the final
    detection performance.  Actually, one can always apply simple
    classifiers to remove a large percentage of negative windows to
    speed up the detection.
}


\subsection{Why LDA Works Better Than LAC}
\label{sec:why}

    Wu et al.\ observed that in many cases,
    LDA post-pro\-c\-e\-ss\-ing gives better
    detection rates on MIT+CMU face data than LAC \cite{wu2008fast}.
    When using the LDA criterion to select Haar features,
    \citet{GSLDA2010Shen} tried different combinations
    of the two classes' covariance matrices
    for calculating the within-class matrix:
    $ \C_w =  \bSigma_1 + \delta \bSigma_2  $ with $ \delta $ being a nonnegative constant.
    It is easy to see that $ \delta = 1 $ and $ \delta = 0 $
    correspond to LDA and  LAC, respectively.
    They found that setting $ \delta \in  [0.5, 1] $
    gives best results on the MIT+CMU face detection task
	\cite{paul2009cvpr,GSLDA2010Shen}.

    According to the analysis in this work, LAC is optimal
    if the distribution of $[h_1( \x )$, $h_2(\x)$, $\cdots$, $h_n(\x) ]$ on
    the negative data  is symmetric. In practice, this requirement may not be
    perfectly satisfied, especially for the first several node classifiers.
    This may explain why in some cases the improvement of LAC is not
    significant.
    However, this does not explain why LDA (FisherBoost) works;
    and sometimes it
    performs even better
    than LAC (LACBoost). At the first glance,
    LDA (or FisherBoost) by no means  explicitly considers the
    imbalanced node learning objective.
    Wu et al. did not have a plausible explanation either
    \cite{wu2008fast,wu2005linear}.

\begin{table*}[t!]
  \centering  
  \begin{tabular}{r|ccccc}
  \hline
        &  $\delta = 0$ (LACBoost) & $\delta = 0.1$ & $\delta = 0.2$ & $\delta = 0.5$ & $\delta = 1$ (FisherBoost) \\
  \hline
  \hline
Digits & $99.12$ ($0.1$) & $\mathbf{99.57}$ ($\mathbf{0.2}$) & $\mathbf{99.57}$ ($\mathbf{0.1}$) & $99.55$ ($0.1$)  & $99.40$ ($0.1$) \\
Faces & $98.63$ ($0.3$) & $98.82$ ($0.3$) & $98.84$ ($0.2$) & $98.48$ ($0.4$) & $\mathbf{98.89}$ ($\mathbf{0.2}$) \\
Cars & $96.80$ ($1.5$) & $97.47$ ($1.1$) & $97.69$ ($1.2$) & $\mathbf{97.96}$ ($\mathbf{1.2}$) & $97.78$ ($1.2$) \\
Pedestrians & $99.12$ ($0.4$) & $\mathbf{99.31}$ ($\mathbf{0.1}$) & $99.22$ ($0.1$) & $99.13$ ($0.3$) & $98.73$ ($0.3$) \\
Scenes & $97.50$ ($1.1$) & $98.30$ ($0.6$) & $98.62$ ($0.7$) & $99.16$ ($0.4$) & $\mathbf{99.66}$ ($\mathbf{0.1}$) \\
\hline
Average (\%) & $98.23$ & $98.69$ & $98.79$ & $98.86$ & $\mathbf{98.89}$ \\
  \hline
  \end{tabular}
  \caption
  {
  The average detection rate and its standard deviation (in \%)
  at $50\%$ false positives.
  We vary the value of $\delta$,
  which balances the ratio between positive and negative class's covariance matrices.
  }
\label{tab:reglac_delta}
\end{table*}

\begin{figure}[t]
    \centering
          \includegraphics[width=.45\textwidth]{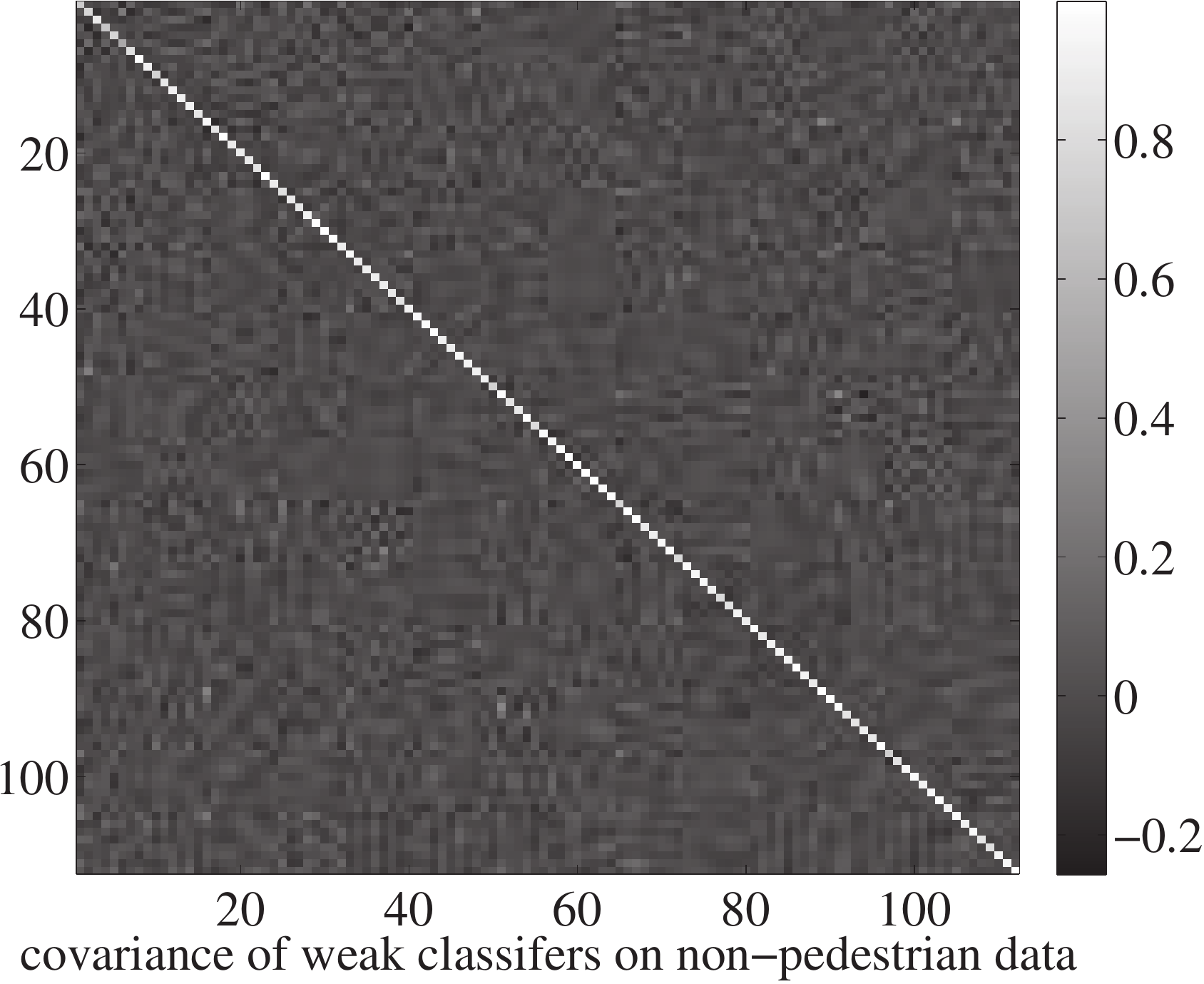}
    \caption{The covariance matrix of the first $112$ weak classifiers selected by
    FisherBoost on non-pedestrian data. It may be approximated by a scaled identity
    matrix.
    On average, the magnitude of diagonal elements is $20$ times larger than those
    off-diagonal elements.
    }
    \label{fig:cov}
\end{figure}

    \begin{proposition}
        For object detection problems,
        the Fish\-er linear discriminant analysis can be viewed as a regularized version
        of linear asymmetric classifier. In other words,
        linear discriminant analysis has already considered
        the asymmetric learning objective.
        In FisherBoost, this regularization is equivalent to
        having a $ \ell_2 $-norm penalty on the primal variable $\w$
        in the objective function of the QP problem in Section \ref{sec:LACBoost}.
        Having the $ \ell_2 $-norm regularization, $ \Vert \w \Vert_2^2 $,
        avoids over-fitting and increases the robustness of FisherBoost.
        This similar pe\-na\-l\-t\-y is also used in machine learning algorithms
        such as Ridge regression (also known as Tikhonov regularization).

    \end{proposition}
        For object detection such as face and pedestrian detection considered here,
        the covariance matrix of the negative class is close to a scaled
        identity matrix.
        In theory, the negative data can be anything other than
        the target. Let us look at one of the off-diagonal elements
        \begin{align}
        \bSigma_{ij, i\neq j}
        &=  {\mathbb E} \bigl[ ( h_i( \x ) - {\mathbb E}[ h_i( \x ) ]  )
                         ( h_j( \x ) - {\mathbb E}[ h_j( \x ) ]  )
                       \bigr]
        \notag
        \\
        &= {\mathbb E} \bigl[  h_i ( \x )  h_j( \x ) \bigr]
                \approx 0.
        \label{EQ:900}
        \end{align}
        Here $ \x $ is the image feature of
        the negative class.
        We can assume that $ \x $ is i.i.d. and
        approximately, $ \x $ follows a symmetric distribution.
        So $ {\mathbb E} [ h_{i,j} ( \x )]   = 0 $.
        That is to say, on the negative class,
        the chance of $ h_{i,j} ( \x ) = +1 $
        or $ h_{i,j} ( \x ) = -1 $ is the same, which is $ 50\% $.
        Note that this does not apply to the positive class
        because $ \x $ of the positive class is not
        symmetrically distributed, in general.
        The last equality of \eqref{EQ:900}
        uses the fact that weak classifiers
        $ h_i( \cdot)  $ and $ h_j( \cdot)  $
        are approximately statistically independent.
        Although this assumption may not hold in practice
        as pointed out in \cite{shen2010dual},
        it could be a plausible approximation.

        Therefore, the off-diagonal elements of  $ \bSigma $
        are almost all zeros; and $\bSigma$ is a diagonal matrix.
        Moreover in object detection, it is a reasonable assumption
        that the diagonal elements
        $ {\mathbb E} [  h_j ( \x )  h_j( \x ) ] $
        $ (j=1,2,\cdots ) $ have similar values.
        Hence, $ \bSigma_2 \approx v \bf I$ holds,
        with $ v $ being a small positive constant.

        So for object detection,
        the only difference between LAC and LDA is that,
        for LAC, $ \C_w = \frac{m_1}{m} \bSigma_1 $
        and for LDA, $ \C_w = \frac{m_1}{m}
                      \bSigma_1
                      + v \cdot \frac{m_2}{m} \bf I $.

    In summary, LDA-like approaches (\eg, LDA post-processing
    and FisherBoost) perform better than LAC-like approaches
    (\eg, LAC and LACBoost) in object detection due to two main reasons.
    The first reason is that LDA is a regularized version of LAC.
    The second reason is that the negative data are not
    necessarily symmetrically distributed. Particularly,
    in latter nodes, bootstrapping forces the negative data
    to be visually similar the positive data.
    In this case, ignoring the negative data's covariance information
    is likely to deteriorate the detection performance.

    Fig.~\ref{fig:cov} shows some empirical evidence
    that $ \bSigma_2 $ is close to a scaled identity matrix.
    As we can see, the diagonal elements are much larger than those
    off-diagonal elements (off-diagonal ones are close to zeros).

In this experiment,
we evaluate the impact of the regularization parameter by  varying
the value of $\delta$,
which balances the ratio between positive and negative class's covariance matrices,
\ie,
$\C_\w = {\bSigma}_1 + \delta {\bSigma}_2$; and also 
$ \Q = \begin{bmatrix} \Q_1 & {\bf 0} \\ {\bf 0} & \delta \Q_2  \end{bmatrix} $.
Setting $\delta = 0$ corresponds to LACBoost,
$ \begin{bmatrix} \Q_1 & {\bf 0} \\ {\bf 0} & {\bf 0}  \end{bmatrix} $,
    while setting $\delta = 1$ corresponds to FisherBoost,
$ \Q = \begin{bmatrix} \Q_1 & {\bf 0} \\ {\bf 0} & \Q_2  \end{bmatrix} $.

We conduct our experiments on $5$ visual data sets by setting the value of
$\delta$ to be $\{0$, $0.1$, $0.2$, $0.5$, $1\}$.
All $5$ classifiers are trained to remove $50\%$ of the negative data, 
while retaining almost all positive data.
We compare their detection rate in Table \ref{tab:reglac_delta}.
First, in general, we observe performance improvement when we set 
$ \delta $ to be a small positive value. 
Since setting $\delta$ to be $1$ happens to coincide with the LDA
objective criterion, the LDA classifier also inherits the node
learning goal of LAC in the context of object detection.
Second, on different datasets, in theory this parameter should be
cross validated and setting it to be $ 1 $ (FisherBoost) does not 
always give the best performance, which is not surprising.

     At this point, a hypothesis  naturally arises: 
     {\em 
     If regularization is really the reason why LACBoost
     underperforms FisherBoost,
     then applying other forms of regularization to LACBoost would also
     be likely to improve LACBoost. 
     }
     Our last experiment tries to verify this hypothesis. 
    
     Here we regularize the matrix $ \Q $ by adding an appropriately scaled identity
     matrix $  \Q + \deltap \bf I  $.   
     As discussed in Section 
     \ref{sec:dual_prob}, from a numerical stability point of view,
     difficulties arise when $ \Q $ is
     rank-deficient, causing  the dual solution to 
     \eqref{EQ:dual} to be non-uniquely defined.
     This issue is much worse for LACBoost because the lower-block
     of $ \Q$, \ie, $ \Q_2 $ is a zero matrix. 
     In that case, a well-defined problem  can be obtained by replacing
     $ \Q $ with $  \Q + \deltap \bf I  $.
     This can be interpreted as
     corresponding to  the
     primal-regularized QP (refer to \eqref{EQ:QP1}): 
      \begin{align}
            \min_{\w,\brho}
             ~&
            \tfrac{1}{2} \brho ^\T \Q \brho - \theta \e^\T
            \brho  +  \deltap \norm{\brho} ^2 ,
          \notag\\
            \quad {\rm s.t.} ~&\w \psd {\bf 0},
             {\bf 1}^\T \w = 1,
      \notag \\
      &
     {\rho}_i = ( \A \w )_i,
        i = 1,\cdots, m.
            \label{EQ:QP1B}
        \end{align}
     Clearly here in the primal, we are applying     
     the Tikhonov $ \ell_2$ norm regularization to the variable $
     \brho$. 
     Also we expect accuracy improvement with  this regularization 
     because the margin variance is minimized by minimizing the $
     \ell_2$ norm of the margin while maximizing the weighted mean of
     the margin, \ie, $ \e^\T \brho $.  
     Thus a better margin distribution may be achieved
     \cite{MDBoost2010Shen,shen2010dual}.

    Now we evaluate the impact of the regularization parameter $
    \deltap $ by running experiments on the same data\-set\-s as in the
    last experiment. 
    We vary the values of $\deltap$ and the results of detection
    accuracy are reported in
    Table \ref{tab:reglac_deltap}. 
    Again, the $5$ classifiers are trained to remove $50\%$ of the
    negative data, while correctly classifying as most  positive data
    as possible.
    As can be seen, indeed, regularization often improves the results.
    Note that in the experiments, we have solved the primal
    optimization problem so
    that even when $ \Q $ is not invertible, we can still obtain a
    solution.  Having the primal solutions,
    the dual solutions are obtained using \eqref{EQ:KAb}. 
    This experiment demonstrates that other formats of regularization
    indeed improves LACBoost too. 

\begin{table*}[t]
\centering
\scalebox{0.95}
{
\begin{tabular}{r|ccccccc}
\hline
        &  $ \deltap = 0 $ (LACBoost) &  $\deltap = 5 \times 10^{-4}$ & $\deltap = 2 \times 10^{-4}$ & $\deltap = 10^{-4}$ & $\deltap = 5 \times 10^{-5}$ & $\deltap = 2 \times 10^{-5}$ & $\deltap = 10^{-5}$ \\
\hline
\hline
Digits & $99.12$ ($0.1$) &  $99.50$ ($0.1$) & $99.41$ ($0.2$) & $99.59$ ($0.2$) &  $\mathbf{99.60}$ ($\mathbf{0.2}$) &  $99.50$ ($0.3$) & $99.11$ ($0.5$) \\
Faces & $98.63$ ($0.3$) &  $98.73$ ($0.0$) & $98.87$ ($0.0$) & $99.02$ ($0.0$) & $98.38$ ($0.0$) & $98.84$ ($0.0$) &  $\mathbf{99.04}$ ($\mathbf{0.0}$) \\
Cars & $\mathbf{96.80}$ ($\mathbf{1.5}$) &  $96.62$ ($1.5$) & $\mathbf{96.80}$ ($\mathbf{1.5}$) & $\mathbf{96.80}$ ($\mathbf{1.5}$) & $96.67$ ($1.4$) & $96.58$ ($1.5$) &  $96.71$ ($1.5$) \\
Pedestrians & $99.12$ ($0.4$) &  $96.62$ ($1.5$) &  $\mathbf{99.32}$ ($\mathbf{0.1}$) &  $99.22$ ($0.2$) &  $98.97$ ($0.4$) &  $98.97$ ($0.3$) &  $98.81$ ($0.4$) \\
Scenes & $97.50$ ($1.1$) &  $\mathbf{98.96}$ ($\mathbf{0.4}$) & $98.36$ ($0.5$) & $97.88$ ($0.6$) & $98.38$ ($0.6$) & $98.25$ ($0.8$) &  $97.1$ ($0.8$) \\
\hline
Average (\%) & $98.23$ & $98.09$ & $\mathbf{98.55}$ & $98.50$ & $98.40$ & $98.43$ &  $98.20$ \\
  \hline
  \end{tabular}
  }
  \caption{
  {
  The average detection rate and its standard deviation (in \%)
  at $50\%$ false positives of various regularized LACBoosts.
  We vary the value of $\deltap$, \ie, $\Q + \deltap {\bf I} $.
  Regularization often improves the overall detection accuracy.
  }}
\label{tab:reglac_deltap}
\end{table*}

\section{Conclusion}
\label{sec:con}

    By explicitly taking into account the node learning goal in cascade classifiers, we have
    designed new boosting algorithms for more effective object detection.

    Experiments validate the
    superiority of the methods developed, which we have labeled FisherBoost and LACBoost.  We have
    also proposed the use of entropic gradient descent to efficiently implement FisherBoost and
    LACBoost. The proposed algorithms are easy to implement and can be applied to other asymmetric
    classification tasks in computer vision.  We aim in future to design new asymmetric boosting
    algorithms by exploiting asymmetric kernel classification methods such as \cite{Tu2010Cost}.
    Compared with stage-wise AdaBoost, which is parameter-free, our boosting algorithms need to tune
    a parameter.

    We are also interested in developing parameter-free
    stage-wise boosting that considers the node learning
    objective.
    Moreover, the developed boosting algorithms only work for
    the case $ \gamma_\circ \leq 0.5$ in \eqref{eq:2}. How can we make it work for
    $ \gamma_\circ \geq 0.5$?
    Last, to relax the symmetric distribution
    requirement for the feature responses of the negative class
    is also a topic of interest.

\small
\bibliographystyle{plainnat}

{
\appendix

\normalsize

\section{Proof of Theorem \ref{thm:2}}

\label{App:MPMa}

   Before we present our results, we introduce an important proposition from 
    \cite{yu2009general}. Note that we have used different notation.
    \begin{proposition}
        For a few different distribution families, the worst-case constraint 
        \begin{equation}
            \label{eq:3}
             \left[  \inf_{ \x\sim ( \bmu, \bSigma )  } \Pr \{ \w^\T \x\leq b \}  \right]
                  \geq \gamma,
        \end{equation}
        can be written as:
        \begin{enumerate}
        \item
            if $ \x\sim (\bmu, \bSigma ) $, {\em i.e.}, 
            $ \x$ follows an arbitrary distribution with 
        mean $ \bmu $ and covariance $ \bSigma $, then
        \begin{equation}
            \label{eq:5A}
                    b \geq 
                        \w ^\T \bmu + \sqrt{ \tfrac{ \gamma }{ 1 - \gamma } }
                        \cdot 
                        \sqrt{  \w^\T \bSigma  \w }; 
        \end{equation}
        \item
            if $ \x\sim (\bmu, \bSigma )_{\rm S},$\footnote{Here
            $(\bmu, \bSigma )_{\rm S}$ denotes
            the family of distributions in $ ( \bmu, \bSigma  ) $ that are also symmetric 
            about the mean $ \bmu $. 
            $(\bmu, \bSigma )_{\rm SU}$ denotes
            the family of distributions in $ ( \bmu, \bSigma  ) $ that are additionally symmetric 
            and linear unimodal about  $ \bmu $.
            }
            then we have
        \begin{equation}
            \label{eq:5B}
            \begin{cases}
                    b \geq 
                        \w ^\T \bmu + \sqrt{ \tfrac{ 1 }{ 2 (1 - \gamma) } }
                        \cdot 
                        \sqrt{  \w^\T \bSigma  \w }, 
                             & \text{if~} \gamma \in (0.5,1);  
                    \\
                        b \geq 
                        \w ^\T \bmu, 
                             & \text{if~} \gamma \in (0,0.5];
            \end{cases}
        \end{equation}
    \item
        if
        $ \x\sim (\bmu, \bSigma )_{\rm SU} $,
        then 
        \begin{equation}
            \label{eq:5C}
            \begin{cases}
                    b \geq 
                        \w ^\T \bmu + \frac{2}{3}  \sqrt{ \tfrac{ 1 }{ 2 (1 - \gamma) } }
                        \cdot 
                        \sqrt{  \w^\T \bSigma  \w }, 
                             & \text{if~} \gamma \in (0.5,1);  
                    \\
                        b \geq 
                        \w ^\T \bmu, 
                             & \text{if~} \gamma \in (0,0.5];
            \end{cases}
        \end{equation}
    \item
        if $ \x$ follows a Gaussian distribution with
        mean $ \bmu $ and covariance  $ \bSigma $, \ie, $ \x\sim {\cal G}( \bmu, \bSigma ) $,
        then
        \begin{equation}
            \label{eq:5D}
                    b \geq 
                    \w ^\T \bmu + \Phi^{-1} ( \gamma ) 
                    \cdot   \sqrt{  \w^\T \bSigma  \w }, 
        \end{equation}
        where $ \Phi(\cdot)$ is the cumulative distribution function (c.d.f.) of the
        standard normal distribution $ {\cal G} (0,1)$, and $ \Phi ^{-1} (\cdot)$
        is the inverse function of $ \Phi(\cdot)$. 
        
        Two useful observations about $ \Phi ^{-1} (\cdot)$ are: 
        $ \Phi ^{-1} ( 0.5) = 0 $; 
        and $ \Phi ^{-1} ( \cdot ) $ is a monotonically increasing 
        function in its domain.  
\end{enumerate}
\label{prop:1}
    \end{proposition}
    We omit the proof of Proposition~\ref{prop:1} here and refer the reader to \cite{yu2009general} for details. 
    Next we begin to prove Theorem~\ref{thm:2}:

    \begin{proof}
        The second constraint of \eqref{eq:2} is simply 
        \begin{equation}
            b \geq \w^\T \bmu_2.
            \label{eq:con1}
        \end{equation}
        The first constraint of \eqref{eq:2} can be handled by writing 
        $ \w^\T \x_1 \geq b $ as $ -  \w^\T \x_1 \leq - b $ 
        and applying the results 
        in Proposition \ref{prop:1}. It can be written as 
        \begin{equation}
            \label{eq:con2}
            - b + \w^\T \bmu_1 \geq \varphi( \gamma ) 
                     \sqrt{ \w^\T \Sigma_1 \w }, 
        \end{equation}
        with \eqref{eq:6}.

        Let us assume that $ \Sigma_1 $ is strictly positive definite 
        (if it is only positive semidefinite, we can always add a small 
        regularization to its diagonal components).
        From \eqref{eq:con2} we have
        \begin{equation}
        \varphi( \gamma ) \leq \frac{ -b + \w^\T \bmu_1 } { \sqrt{ \w^\T \Sigma_1 \w }}.
        \label{eq:11}
        \end{equation}
        So the optimization problem becomes 
        \begin{equation}
        \max_{\w, b, \gamma } \, \gamma, \,\, 
                \st \,\, \eqref{eq:con1} \text{~and~}
        \eqref{eq:11}.
        \end{equation}
        
        The maximum value of $\gamma$ (which we label $ \gamma^\star $)
        is achieved when  \eqref{eq:11}
        is strictly an equality. 
        To illustrate this point,
        let  us assume that the maximum is achieved when 
        \[
        \varphi( \gamma^\star ) < \frac{ -b + \w^\T \bmu_1 } { \sqrt{ \w^\T \Sigma_1 \w }}.
        \] 
        Then a new solution can be obtained
        by increasing $  \gamma^\star $ with a positive value such that 
        \eqref{eq:11} becomes an equality. Notice that the constraint \eqref{eq:con1}
        will not be affected,
        and the new solution will be better than the previous one.
        Hence, at the optimum, \eqref{eq:opt} must be fulfilled. 

        Because $ \varphi (\gamma)  $ is monotonically increasing 
        {\it for all the four cases}
        in its domain $ (0,1) $ (see Fig.~\ref{fig:1}),
        maximizing $ \gamma $ is equivalent to maximizing 
        $\varphi (\gamma)  $ 
        and this results in 
        \begin{equation}
            \label{eq:11b}
            \max_{\w, b } \,
            \frac{ -b + \w^\T \bmu_1 } 
                { \sqrt{ \w^\T \Sigma_1 \w } }, \,\, 
                \st \,
                b \geq \w^\T \bmu_2.    
        \end{equation}
        As in \cite{lanckriet2002mpm,huang2004mpm}, we also have a scale ambiguity: 
        if $ ( \w^\star, b^\star )$ is a solution,
        $ (  t \w^\star,  t b^\star ) $ with $ t > 0 $ is also 
        a solution.
        
        An important observation is that  the problem \eqref{eq:11b}
        must attain the optimum at \eqref{EQ:11c}.
        Otherwise if $ b >  \w^\T \bmu_2$, the optimal value
        of \eqref{eq:11b} must be smaller. 
        So we can rewrite \eqref{eq:11b} as an unconstrained problem
        \eqref{eq:11d}.

        We have thus shown that, if $ \x_1 $ is distributed
        according to a symmetric, symmetric unimodal, or Gaussian distribution,
        the resulting optimization problem is identical. This is not surprising 
        considering the latter two cases 
        are merely special cases of the symmetric distribution family. 

        At optimality, the inequality \eqref{eq:11} becomes an
        equality, and hence
        $ \gamma^\star$ can be obtained as in \eqref{eq:opt}.
        For ease of exposition, let us denote the fours cases in the right side of 
        \eqref{eq:6} as
        $ \varphi_{\rm gnrl} ( \cdot)$,
        $ \varphi_{\rm S} ( \cdot) $, 
        $ \varphi_{\rm SU} ( \cdot) $, and
        $ \varphi_{\cal G} ( \cdot) $. 
        For $ \gamma \in [0.5, 1) $, as shown in Fig.~\ref{fig:1},
        we have
            $  
                  \varphi_{\rm gnrl} ( \gamma )
               >  \varphi_{\rm S}  ( \gamma ) 
               >  \varphi_{\rm SU} ( \gamma )
               >  \varphi_{\cal G} ( \gamma ) 
               $. 
        Therefore, when solving \eqref{eq:opt} for $ \gamma^\star $, we have 
        $ 
          \gamma^\star_{\rm gnrl}  <
          \gamma^\star_{\rm S}  <      
          \gamma^\star_{\rm SU} <
          \gamma^\star_{\cal G}$.
        That is to say, one can get better accuracy when
        additional information about the data distribution is available, although
        the actual optimization problem to be solved is identical.
    \end{proof}
    
    \begin{figure}[t]
        \begin{center}
            \includegraphics[width=1\linewidth]{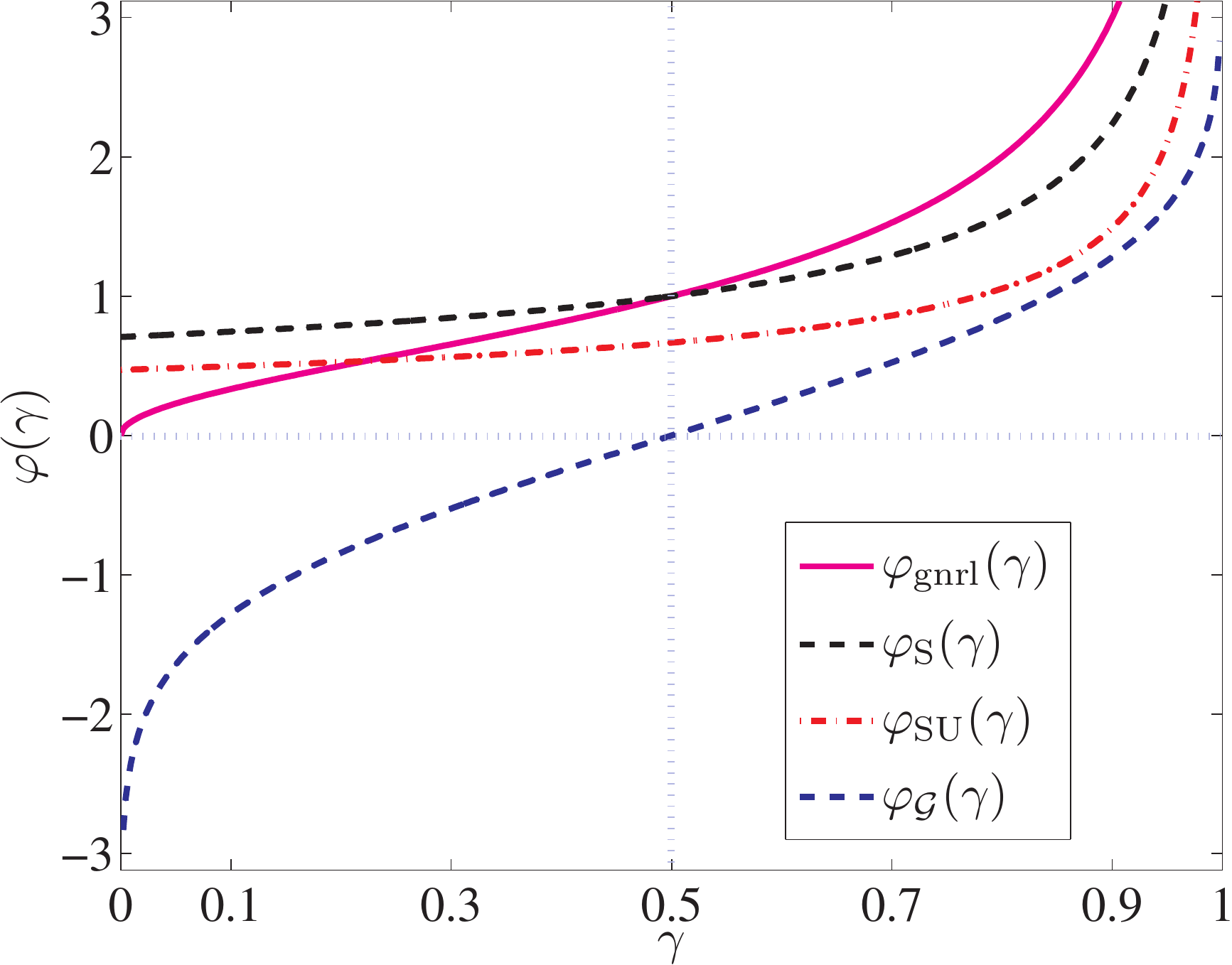}
        \end{center}
        \caption{The function $ \varphi(\cdot) $ in \eqref{eq:6}. The four curves correspond to
        the four cases. They are all monotonically increasing in $(0,1)$.}
        \label{fig:1}
    \end{figure}


\section{Proof of Theorem \ref{thm:CG}}

\label{App:CG}

    Let us assume  that in the current solution we have 
    selected $ n $ weak classifiers and their
    corresponding linear weights are $ \w = [ w_1, \cdots, w_n ] $.
    If we add a weak classifier $
    h'(\cdot) $  that is not in the current subset, the corresponding $ w $ is zero,
    then we can conclude that the current weak classifiers and $\w$
    are the optimal solution already.
    In this case, the best weak classifier that is found by solving the subproblem 
    \eqref{EQ:pickweak} 
    does not contribute to solving the master problem. 

    Let us consider the case that the optimality condition is violated.
    We need to show that we are able to find such a weak learner $ h'(\cdot) $,
    which is not in the set of current selected weak
    classifiers, that its corresponding coefficient $ w > 0 $ holds.
    Again assume $ h'(\cdot) $ is the most violated weak learner 
    found by solving  \eqref{EQ:pickweak} and the convergence condition 
    is not satisfied. In other words, we have 
    \begin{equation}
        \label{EQ:APPCG1}
        \sum_{i=1}^m u_i y_i h'( \x_i ) \geq r. 
    \end{equation}

    Now, after this weak learner is added into the master problem, the corresponding primal solution
    $ w $ must be non-zero (positive because we have the nonnegative-ness constraint on $ \w $).
    
    If this is not the case, then the corresponding $ w = 0 $. 
    This is not possible because of the following reason. 
    From the Lagrangian  
    \eqref{EQ:Lag1}, at optimality we have $ \partial L / \partial \w = {\boldsymbol  0} $,
    which leads to 
    \begin{equation}
        \label{EQ:APPCG2}
         r - \sum_{i=1}^m u_i y_i h'( \x_i ) = q > 0.
    \end{equation}
    Clearly \eqref{EQ:APPCG1}
    and  \eqref{EQ:APPCG2} contradict. 

    Thus, after the weak classifier $ h'(\cdot) $ is added to the
primal problem, its corresponding $ w $ must have a positive
solution. This is to say,  one more free variable is added into
the problem and re-solving the primal problem \eqref{EQ:QP1} must reduce
the objective value. Therefore a strict decrease in the objective is
obtained. In other words, Algorithm \ref{alg:QPCG}
must make progress at each iteration.
    Furthermore, the primal optimization problem is convex, there are no local optimal points. The
    column generation procedure is guaranteed to converge to the global optimum up to some
    prescribed accuracy.

\section{Exponentiated Gradient Descent} 

\label{App:EG}

    Exponentiated Gradient Descent (EG) is a very useful tool for solving large-scale 
    convex minimization problems over the unit simplex. 
    Let us first define the unit simplex 
    $ \Delta_n =  \{ 
    \w \in \Real^n  :  {\bf 1 } ^ \T \w = 1, \w \psd {\bf 0 }
    \} $. 
    EG efficiently solves the convex optimization problem
    \begin{equation}
        \label{EQ:EG1}
        \min_\w \,\,\, f(\w), \,
        {\rm s.t.} \,\, \w \in \Delta_n,  
    \end{equation}
    under the assumption that the objective function $ f(\cdot) $
    is a convex Lipschitz continuous function with Lipschitz
    constant $ L_f $ w.r.t. a fixed given norm $ \lVert \cdot \rVert$.
    The mathematical definition of $ L_f $ is that
    $  | f(\w) -f (\z) |  \leq L_f   \lVert  \x - \z  \rVert$ holds
    for any $ \x, \z $ in the domain of $ f(\cdot)$.
    The EG algorithm is very simple:
    \begin{enumerate}
    \item
        Initialize with $\w^0 \in \text{the interior of }  \Delta_n$;
    \item
        Generate the sequence $ \{ \w^k \} $, $ k=1,2,\cdots$
        with:
        \begin{equation}
            \label{EQ:EQ2}
            \w^k_j = \frac{ \w^{k-1}_j \exp [ - \tau_k f'_j ( \w^{k-1} ) ]  } 
            { \sum_{j=1}^n  \w^{k-1}_j \exp [ - \tau_k f'_j ( \w^{k-1} ) ] }. 
        \end{equation}
        Here $ \tau_k $ is the step-size. 
        $ f'( \w ) = [ f_1'(\w), \dots,  f_n'(\w) ] ^\T $
        is the gradient of $ f(\cdot) $;
    \item
        Stop if some stopping criteria are met.
    \end{enumerate}
    The learning step-size can be determined by 
        \begin{equation*}
            \tau_k = \frac{ \sqrt{ 2\log n } } { L_f }
                     \frac{1}{ \sqrt{ k } },
        \end{equation*} 
    following \cite{beck03mirror}.
    In \cite{globerson07exp}, the authors have 
    used a simpler strategy to set the learning rate. 

    In EG there is an important parameter $ L_f $, which is
    used to determine the step-size. 
    $ L_f $ can be determined by the $\ell_\infty $-norm of $ | f' (\w) | $.
    In our case $ f' (\w) $ is a linear function, which is trivial to compute.
    The convergence of EG is guaranteed; see \cite{beck03mirror} for details.

}

\end{document}